\documentclass[twocolumn,10pt]{IEEEtran}

\usepackage{rotating}
\usepackage{upgreek}
\usepackage{array}
\usepackage{multirow}
\usepackage{float}

\usepackage[utf8]{inputenc}
\usepackage[T1]{fontenc}

\usepackage[colorlinks=false]{hyperref}

\usepackage{amsmath, amsfonts,amsthm,amssymb}
\usepackage{bbm}

\usepackage{graphicx}
\usepackage{epstopdf}

\usepackage{txfonts}

\usepackage[font=footnotesize,labelfont=bf]{caption}

\usepackage{color}

\usepackage{customCommands}

\theoremstyle{remark}

\usepackage{aliascnt}
\newaliascnt{conj}{theorem}
\newaliascnt{cor}{theorem}
\newaliascnt{lemma}{theorem}
\newaliascnt{prop}{theorem}
\newaliascnt{definition}{theorem}
\newaliascnt{example}{theorem}
\newaliascnt{notation}{theorem}
\newaliascnt{experiment}{theorem}

\theoremstyle{theorem}

\theoremstyle{definition}

\usepackage[sort&compress, numbers]{natbib}
\usepackage{snapshot}

\usepackage{booktabs}
\usepackage{multicol}

\usepackage[boxed, ruled, lined]{algorithm2e}

\makeatletter
\def\hlinewd#1{%
\noalign{\ifnum0=`}\fi\hrule \@height #1 %
\futurelet\reserved@a\@xhline}
\makeatother

\usepackage[usenames,dvipsnames,svgnames]{xcolor}

\usepackage{blindtext}

\usepackage{enumerate}
\usepackage{csquotes}

\usepackage[font=footnotesize]{subcaption}

\title{
Fast Piecewise-Affine Motion Estimation Without Segmentation
}

\author{Denis Fortun,
        Martin Storath,
        Dennis Rickert,
        Andreas Weinmann,
        and Michael Unser
\thanks{D. Fortun is with the Signal Processing core of Center for Biomedical Imaging (CIBM-SP), EPFL, Lausanne, Switzerland.}
\thanks{M. Storath is with the  Image  Analysis  and  Learning  Group,  Universit\"at Heidelberg, Germany.}
\thanks{D. Rickert is with the Institute of Computational Biology, Helmholtz Zentrum München, Germany}%
\thanks{A. Weinmann is with the Institute of Computational Biology, Helmholtz Zentrum München, and with the Department of Mathematics and Natural Sciences, Hochschule Darmstadt, Germany.} 
\thanks{M. Unser is with Biomedical Imaging Group, EPFL, Lausanne, Switzerland.}
}

\begin{document}

\maketitle
\IEEEpeerreviewmaketitle

\newlength\figureheight
\newlength\figurewidth
\setlength\figureheight{0.15\textwidth}


\begin{abstract}

Current algorithmic approaches for piecewise affine motion estimation are based on alternating motion segmentation and estimation. We propose a new method to estimate piecewise affine motion fields directly without intermediate segmentation. To this end, we reformulate the problem by imposing piecewise constancy of the parameter field, and derive a specific proximal splitting optimization scheme. A key component of our framework is  an efficient one-dimensional piecewise-affine estimator for vector-valued signals. The first advantage of our approach over segmentation-based methods is its absence of initialization. The second advantage is its lower computational cost which is independent of the complexity of the motion field. In addition to these features, we demonstrate competitive accuracy with other piecewise-parametric methods on standard evaluation benchmarks. Our new regularization scheme also outperforms the more standard use of total variation and total generalized variation.


\end{abstract}

\maketitle

\section{Introduction}

Two important prior models have been explored for motion estimation. The first one works with a dense representation of motion and imposes at each pixel a
smoothness constraint \cite{Horn81} such as total variation (TV) \cite{Brox04,Zach07}. Its regularization terms  are most often convex and well suited to a large collection of
optimization techniques.
The second type of prior model works with parametric representations of motion, which may be chosen to provide a satisfying match of 3D translations on the camera
plane.
In particular, piecewise-parametric estimation methods have yielded very
accurate results \cite{Unger12,Sun10b,Yang15}. Yet, in spite of these achievements, local smoothness priors are still
preferred to piecewise-parametric ones in most motion estimation methods.
The main reason is the difficulty of the optimization problem associated with the piecewise-parametric approach.
In this work, we address this optimization issue.

The problem is usually formulated as the joint segmentation of the motion field and estimation of the parameters inside each region, following the seminal work of Mumford
and Shah \cite{mumford1989optimal} for the image segmentation part. The interdependency of these two tasks translates into highly non-convex
optimization.
The existing solutions proceed iteratively by alternating an optimization step with respect to the image partition and an optimization step with respect ot the motion parameters. This alternance
causes two main issues.
Firstly, the resulting scheme is very sensitive to initialization and can only be used for refinement.
Secondly, the computational cost is often prohibitive for practical applications. In particular, it  depends on the number of regions, which should typically be
very large to achieve high-accuracy. 

In this paper, we propose a new method to estimate piecewise-affine motion fields. 
It eschews the explicit segmentation of  motion, leading to the direct estimation of a piecewise-affine motion field. We revisit the standard formulation and impose a
piecewise-constant regularization of the field of affine parameters.
The key step of our algorithm is a specific proximal-splitting approach that yields a series of 1D piecewise-affine vectorial estimation problems. We propose an efficient
solver that is based on dynamic programming and inspired by the works on segmentation described in \cite{storath2014fast,storath2014joint,weinmann2015iterative}.

Extensive experiments on the reference benchmarks {\it MPI Sintel} \cite{Butler12} and {\it Kitti} \cite{Geiger12} show that our approach outperforms the standard TV and
total generalized variation (TGV) regularizations, and that it is competitive with the best performing piecewise-parametric methods. Moreover, our optimization does not
require any initialization of the motion field, and it is faster than other piecewise-parametric approaches.
In particular, the computational time does not depend on the complexity of the motion field. Thus, our method combines the advantages of the piecewise-affine model with
robustness and a low computational cost. It can be integrated as a regularizer in various motion estimation frameworks.

The outline of the paper is as follows: In Section~\ref{sec:related_works}, we review related works and identify their limitations. In Section~
\ref{sec:pwaff_of}, we describe the model and optimization method that we propose. In Section~\ref{sec:experiments}, we evaluate the performance of our
method on standard benchmarks.

\section{Related Works}
\label{sec:related_works}

Joint motion segmentation and parameter estimation has been formulated as an optimization problem of the form
\begin{equation}
J=\min_{\Rc, \w}~ \left(\rho(\w) + \frac{\lambda}{2} \sum_{n=1}^L \Lc( R_n)\right),
\label{eq:pwaff_mumsha}
\end{equation}
 where $L$ is the number of regions, $\Rc = \{R_1, \ldots, R_L\}$ is a partition of the image, $\w$ is a piecewise-parametric motion field on $\Rc$ (i.e. $\w$ is
 parametric on each $R_n$ for $n=1,\ldots,L.$), $\rho$ is a function that imposes data fidelity, $\Lc$ is a segmentation prior usually
 defined as the total boundary length delineating the segmented regions $R_n,$ and $\lambda$ is a balance parameter between $\rho$ and $\mathcal L$. Problems of this type
 are typically minimized alternately with respect to $\Rc$, which amounts to an image-partitioning problem, and with respect to $\w$, which amounts to a parametric motion  fitting.

The differences between existing methods concern mainly the solver for the image-partitioning problem.
In a continuous setting, following the approach of Mumford and Shah \cite{mumford1989optimal}, the problem has been addressed with an implicit level-set representation of
the partitioning curve in \cite{Cremers05,Paragios05,Vazquez06}. A primal-dual optimization strategy was used in \cite{Unger12}.  
In a discrete setting, iterated conditional modes and high confidence first approaches were exploited in \cite{Bouthemy93,Odobez98,Memin02}. Graph-cuts
methods have also been used in \cite{Schoenemann06}, and more recently in \cite{Yang15}.
Layered models, introduced in \cite{Wang94}, involve a similar optimization problem but add a depth information between the different regions, from which
occlusions can be derived. This model has been revitalized in \cite{Sun10b,Sun12,Sun14,Sevilla16}.

The importance of initialization when optimizing (\ref{eq:pwaff_mumsha}) with an alternating scheme is illustrated in \cite{Sun12,Unger12}, where the optimization is 
initialized through advanced motion estimation methods \cite{Sun10,Werlberger09}. In \cite{Cai14}, an alternating direction method of multipliers (ADMM) approach is used
to solve (\ref{eq:pwaff_mumsha}) without intermediate segmentation steps.
However, the underlying model is piecewise-constant and not rich enough in most practical scenarios; it is initialized by a block matching algorithm.

Most of the computational effort is spent on the image-partitioning
problem.
The earliest works retain at most five regions to make the problem tractable \citep{Bouthemy93,Memin02,Cremers05,Paragios05}.
More recently, the layered approach \citep{Sun12} handles a larger number of regions but requires several hours of computation, and the primal dual approach
\citep{Unger12} can take up to one hour despite a GPU implementation.
The method proposed  in \cite{Yang15} achieves around fifteen minutes for $(1240\times 370)$ image, with a graph cut minimization approach.

Beyond solving (\ref{eq:pwaff_mumsha}), other techniques can be involved to improve the results. They include the handling of occlusions
\cite{Bleyer06,Odobez98,Unger12}, label cost terms to limit the number of regions \cite{Unger12,Yang15}, edge-driven models to fit image boundaries \cite{Paragios05},
 deviations from the parametric models to estimate more complex deformations \cite{Sun10b},  smoothness of the parameters of neighboring regions
\cite{Yang15}, or post-processing refinements with a variational optimization of TV-based models \cite{Yang15}.
Yet other methods rely on similar principles but incorporate additional information obtained from their applicative context, like epipolar constraints
\cite{Hur16,Vogel15}, temporal consistency \cite{Hur16}, or semantic information about the type of moving objects in the scene \cite{Sevilla16}.

Extensions of TV to second order derivatives result in approximately piecewise-affine solutions  \cite{Trobin08a,Ranftl14}.
However, the $\ell_1$ norm does not delineate moving objects as sharply as the Mumford-Shah model (\ref{eq:pwaff_mumsha}). In this line, the over-parametrized approach
\cite{Nir08,Hornacek14}, which models a spatially varying parameter field with TV regularization, also shows this undesirable effect.

\section{Proposed Piecewise-Affine Estimation}
\label{sec:pwaff_of}
In this section, we detail our method to estimate piecewise-affine motion fields. After the model and minimization
problem, we present our optimization strategy based on directional splitting. The key to our method is an efficient solver for the vectorial 1D
piecewise-affine denoising problem.

\subsection{Piecewise-Affine Model}

Let two successive frames of an image sequence be $I_1,I_2:\Omega\rightarrow\mathbb R$, where $\Omega\subset\mathbb N^2$ is the  image grid. 
Our goal is to estimate the piecewise-affine motion field
$\w:\Omega\rightarrow\mathbb R^2$ that transports $I_1$ to $I_2$
according to \eqref{eq:pwaff_mumsha}.
It is common to discretize the length $\Lc$ of a segment boundary 
by 
\begin{equation}
    \Lc(R_n) = \sum_{k=1}^K \alpha_k | \{\x \in R_n :  \x + \d_k \notin R_n \}|,
\end{equation}
where  $\d_k$ is an element of the  set $\mathcal D  \subset \mathbb (Z^2)^K$ of directions 
 and  $\alpha_k > 0$ is its corresponding weight \cite{blake1987visual,chambolle1999finite}.
The choice of $\mathcal D$ and $\alpha$ determines how well the regularizer approximates rotational invariance.
Considering only horizontal and vertical directions, with $\mathcal D=\{(0,1),(1,0)\}$, creates block artifacts similar to those of anisotropic TV
regularization. To attenuate them, we use the four-directional neighborhood system $\mathcal D=\{(0,1),(1,0),(1,1),(-1,1)\}$ that includes diagonal
directions. The weights are chosen such that the norm built from the basis vectors of $\mathcal D$ best approximates the isotropic Euclidean
norm \cite{chambolle1999finite,storath2014joint}.

Henceforth, we assume that the motion field $\w$
can be written in terms of the parameter field $\P: \Omega \to \R^{2\times 3}$ as
\begin{equation}
    \w(\x) = \P(\x)\bar{\x}
\end{equation}
$\text{for all }\x \in \Omega$, where $\bar{\x} = (x_1,x_2,1)$ denotes the homogeneous coordinates of $\x.$
When $\P$ is piecewise constant, it defines a partition $\mathcal \Rc$
of the domain $\Omega$.
This allows us to conveniently express
the piecewise-affine model \eqref{eq:pwaff_mumsha} as
\begin{equation}\label{eq:ms_para_field}
\begin{split}
J=&\min_{\w,\P}~\left(\rho(\w) + \lambda \sum_{k=1}^K\alpha_k   
\| \mbox{\boldmath{$\mbox{\boldmath{$\nabla$}}$}}_{\d_k} \P \|_0\right) \\
&\text{s.t. }\w(\x) = \P(\x)\bar{\x} 
\end{split}
\end{equation}
$\text{for all }\x \in \Omega$, where $\| \mbox{\boldmath{$\nabla$}}_{\d_k} \P \|_0$ counts the number of parameter changes with respect to the direction $\d_k$,
as given by
\begin{equation}
   \| \mbox{\boldmath{$\nabla$}}_{\d_k} \P \|_0  = 
   \left|\{ \x \in\Omega : \P({\bf x})\neq \P({\bf x}+{\bf d}_k) \}\right|.
\end{equation}
Note that the factor $\frac{1}{2}$ 
that was compensating for the double counting of the boundary lengths in \eqref{eq:pwaff_mumsha}
is not needed in \eqref{eq:ms_para_field}.
Although our final goal is to estimate the flow field $\w$, the introduction of $\P$ in
\eqref{eq:ms_para_field} is important for the derivation of our proposed algorithm. Differently from the over-parametrized approach \cite{Nir08,Giryes15}, we do not estimate the parameters but
directly the motion field.

The data term in (\ref{eq:pwaff_mumsha}) reflects the assumption of the conservation of an image feature along the motion trajectory. Here, we rely on the usual
assumption of constant brightness and penalize deviations  with an $\ell_1$ norm to gain robustness to local violations such as occlusions or illumination
changes.
The linearized form of this criterion is
\begin{equation}
\rho_\text{d}(\w) = \sum_{\x\in\Omega}|\mbox{\boldmath{$\nabla$}}^\top I_2(\x)\, \w(\x) + I_t(\x)|,
\label{eq:data_term}
\end{equation}
where $\mbox{\boldmath{$\nabla$}}^\top=\left(\frac{\partial}{\partial x_1},\frac{\partial}{\partial x_2}\right)^\top$ and $I_t$ is the discrete temporal image gradient
given by $I_t=(I_2-I_1)$.
Note that this data term (\ref{eq:data_term}) does not depend on the parameter field $\P$ but only on the associated flow field $\w.$

\subsection{Splitting Approach and Augmented Lagrangian Resolution}
The problem  (\ref{eq:ms_para_field}) is non-convex and NP-hard. Thus, the convergence to a global minimum cannot be guaranteed. To find a practical solution, we devise
a splitting strategy. We divide (\ref{eq:ms_para_field}) into easier subproblems in an ADMM-like augmented-Lagrangian framework, which has turned out to often work well
for non-convex problems \cite{chartrand2013nonconvex, storath2013jump,wang2015global,hohm2015algorithmic,xu2016empirical}.

The starting point for our method is the formulation in terms of the parameter field \eqref{eq:ms_para_field}. 
We introduce splitting variables $\P_k$ to decouple the data term and the terms associated to the directions $\d_k$ of the
regularization. This leads to the reformulation of \eqref{eq:ms_para_field} as
\begin{eqnarray}
 \label{eq:constrained_min}
J&=&\min_{\w, \P_1, \ldots, \P_K} \rho(\w) + \lambda\sum_{k=1}^K \alpha_k \| \mbox{\boldmath{$\nabla$}}_{d_k} \P_k \|_0 \label{eq:constrained_form_2}\\
&& \text{ s.t. }
 \w(\x)=\z_k(\x),\nonumber \\
 &\,& \phantom{\text{ s.t. }}
 \z_k(\x)=\P_k(\x) \bar\x ,~ \forall \x\in\Omega,~ \forall  k = 1,\ldots,K\nonumber.
\end{eqnarray}

Then, the augmented Lagrangian (in scaled form) of (\ref{eq:constrained_min}) writes
\begin{equation}
    \begin{split}
\Ac_\eta&(\w,\{\P_k\}_k,\{\z_k\}_k,\{\mbox{\boldmath{$\upmu$}}_k\}_k) =  \rho(\w) + \lambda\sum_{k=1}^K \alpha_k \| \mbox{\boldmath{$\nabla$}}_{\d_k} \P_k \|_0 \\
&+ \frac{\eta}{2}\sum_{k=1}^K\sum_{\x\in\Omega}\left\|\w(\x)-\z_k(\x)+\frac{\mbox{\boldmath{$\upmu$}}_k(\x)}{\eta}\right\|_2^2 - \frac{1}{2\eta}\left\|\mbox{\boldmath{$\upmu$}}_k(\x)\right\|_2^2, \nonumber \\[1ex]
\text{ s.t. }& 
 \z_k(\x)=\P_k(\x) \bar\x ,~ \forall \x\in\Omega,~ \forall  k\in\{1,\ldots,K\},
 \end{split}
 \label{eq:AL}
\end{equation}
where $\{\mbox{\boldmath{$\upmu$}}_k\}_{k=1,\ldots,K}$ are Lagrange multipliers and $\eta > 0$ is a parameter that controls the fulfillment of the  constraints, and
influences the speed of convergence.
Further, $\|\cdot \|_2^2$ denotes the squared Euclidean norm in $\R^2$.
Note that we include the equality constraints  into the target functional only 
with respect to the motion field variables. 
The couplings of the parameter fields $\P_k$ and the flow fields
$\z_k$ remain as explicit constraints. 
This will become important when solving the subproblems.

Next, we follow the ADMM strategy and iteratively minimize the augmented Lagrangian with respect to $\w$ and $\z_k$, and
perform gradient ascents on the Lagrange multipliers as
\begin{eqnarray}
&\,&\w^{(n+1)}=\argmin_\w \Ac_\eta(\w,\z_1^{(n)},\ldots,\z_K^{(n)})  \nonumber\\
&\,& (\z^{(n+1)}_1, \cdot)=\argmin_{\z_1, \P_1} \Ac_\eta(\w^{(n+1)},\z_1,\ldots,\z_K^{(n)}, P_1) \nonumber\\
&\,& ~~~~~~~~~~~~~~~~~~~~~~~~~~\text{ s.t. } \z_1(\x) = \P_1(\x)\bar\x,\, \forall \x\in\Omega \nonumber\\
&\,& ~~~~~~~~~~~~~~~~\vdots \nonumber\\
&\,& (\z^{(n+1)}_K , \cdot)=\argmin_{\z_K, \P_K} \Ac_\eta(\w^{(n+1)},\z_1^{(n+1)},\ldots,\z_K, \P_K) \label{eq:admm}
\\
&\,& ~~~~~~~~~~~~~~~~~~~~~~~~~~\text{ s.t. } \z_K(\x) = \P_K(\x),\, \forall \x\in\Omega \nonumber\\[1ex] 
&\,&\mbox{\boldmath{$\upmu$}}_1^{(n+1)}(\x) = \mbox{\boldmath{$\upmu$}}_1^{(i)}(\x) ~+~ \eta(\w^{(n+1)}(\x)-\z_1^{(n+1)}(\x)),\,\forall \x\in\Omega \nonumber\\
&\,& ~~~~~~~~~~~~~~~~\vdots \nonumber\\
&\,&\mbox{\boldmath{$\upmu$}}_K^{(n+1)}(\x) = \mbox{\boldmath{$\upmu$}}_K^{(n)}(\x) ~+~ \eta(\w^{(n+1)}(\x)-\z_K^{(n+1)}(\x)),\,\forall \x\in\Omega. \nonumber
\end{eqnarray}
Observe that we only need the minimizing arguments
with respect to the $\z_k$ variables, but not with respect to the the parameter field.
This is the reason why we omit the minimizer with respect to $\P_k$ on the left hand side of (\ref{eq:admm}).

The accuracy and efficiency of our approach is based on our ability to solve exactly and at low computational cost each  subproblem in (\ref{eq:admm}). In 
Sections \ref{sec:subprb_w} and \ref{sec:subpb_z}, we detail our solvers for the update of $\w$ and $\z_k$, respectively. 

\subsection{Update of $\w$}
\label{sec:subprb_w}
The minimization with respect to $\w$ in (\ref{eq:admm}) writes
\begin{eqnarray}
\hat \w=\argmin_\w \left(\rho(\w) + \frac{\eta}{2}\sum_{k=1}^K\sum_{\x\in\Omega}\left\|\w(\x)- \z_k(\x)+\frac{\mbox{\boldmath{$\upmu$}}_k(\x)}{\eta}\right\|_2^2\right).
\label{eq:min_w}
\end{eqnarray}
With simple manipulations, we  rewrite (\ref{eq:min_w}) as
\begin{eqnarray}
\hat \w=\argmin_\w \left( \rho(\w) ~+~ \frac{\eta K}{2}\sum_{\x\in\Omega}\left\|\w(\x)-\r(\x)\right\|^2\right),
\label{eq:min_w_rewritten}
\end{eqnarray}
where
\begin{eqnarray}
\r(\x) = \frac{1}{K}\sum_{k=1}^K\left(\z_k(\x)-\frac{\mbox{\boldmath{$\upmu$}}_k(\x)}{\eta}\right).
\end{eqnarray}
Problem (\ref{eq:min_w_rewritten}) is pointwise and admits a closed-form solution with the thresholding scheme
  \begin{equation}
  \hat \w(\x) = \r(\x) + 
  \begin{cases}
  \frac{\mbox{\boldmath{$\nabla$}} I}{\eta K}, &\rho_0(\r(\x))<-\frac{\|\mbox{\boldmath{$\nabla$}} I\|_2^2}{\eta K}\\
  -\frac{\mbox{\boldmath{$\nabla$}} I}{\eta K}, & \rho_0(\r(\x))>\frac{\|\mbox{\boldmath{$\nabla$}} I\|_2^2}{\eta K}\\
  -\rho_0(\r(\x))\frac{\mbox{\boldmath{$\nabla$}} I}{\|\mbox{\boldmath{$\nabla$}} I\|_2^2}, & |\rho_0(\r(\x))|\leq\frac{\|\mbox{\boldmath{$\nabla$}} I\|_2^2}{\eta K},
  \end{cases}
  \end{equation}
where $\mbox{\boldmath{$\nabla$}} I=\mbox{\boldmath{$\nabla$}}^\top I_2(\x)$ and $\rho_0(\r(\x)) = |\mbox{\boldmath{$\nabla$}} I\, \r(\x) + I_t|$. A similar step appears in the context of a primal-dual optimization framework~\cite{Zach07,Chambolle11}.

Note that while we give here the solution for a data term derived from  brightness constancy, the pointwise nature of the problem makes it
tractable for other assumptions. For example, solutions for  more sophisticated data fidelity terms based on normalized cross correlation or census
transform are studied in \cite{Vogel13}.

\subsection{Fast update of $\z_k$}
\label{sec:subpb_z}

We address the minimization of the augmented Lagrangian with respect to $\z_k$  in the ADMM steps \eqref{eq:admm}.
It is instructive to first consider the case $\z_1$ which corresponds to the minimization in the vertical direction $\d_1 = (0,1).$


Our first step is to reduce the problem to a one-dimensional parameter estimation.
To this end, we write the corresponding line in \eqref{eq:admm} as
\begin{equation}\label{eq:minZ1}
\begin{split}    
   (\hat \z_1,\cdot) =& \argmin_{\z, \P}\left( \kappa \|\mbox{\boldmath{$\nabla$}}_{\d_1} \P \|_0 
+  \sum_{\x\in\Omega}\left\|\v(\x)-\z(\x) \right\|_2^2\right), \\
&\text{s.t. } \P(\x)\bar \x = \z(\x),
\end{split}
\end{equation}
with $\v(\x) = \w(\x) + \frac{\mbox{\boldmath{$\upmu$}}_k(\x)}{\eta}$
and $\kappa = \frac{2\alpha_k \lambda}{\eta}.$
Recall that $\x \in \{1, \ldots, m\} \times \{1, \ldots, n\} $
and that
$\z \in \R^{2}$ and $\P \in  \R^{2 \times 3}.$
A crucial observation is 
that $\|\mbox{\boldmath{$\nabla$}}_{\d_1} \P \|_0 $ only takes into account 
neighborhood differences  within the vertical scan lines.
Therefore, the two-dimensional optimization problem \eqref{eq:minZ1}
boils down to independent one-dimensional subproblems.
Let us fix a vertical scan line by choosing a fixed index $x_1.$
The $x_1$th line of a  minimizer  $\hat \z_1$  is then given by
\begin{equation}\label{eq:horizontalProblemNew}
\begin{split}
&(\hat \z_1(x_1, \cdot), \cdot) = \argmin_{\z', \P'} \left( \kappa \|\mbox{\boldmath{$\nabla$}}_{\d_1} \P' \|_0 
+ \sum_{m=1}^n \left\| \v'(m) - \z'(m) \right\|^2_2\right), \\
&~~~~\text{s.t. } P'_{11}(m)x_1 + P'_{12}(m)m + P'_{13}(m)  = z'_1(m) \\
&~~~~\phantom{\text{s.t. }} P'_{21}(m)x_1 + P'_{22}(m)m + P'_{23}(m)  = z'_2(m), 
\end{split}
\end{equation}
$\text{for all }m$, where $\v'(m) = \v(x_1, m),$ and where $\z'$ and $\P'$ are the flow field and parameter field on a one-dimensional line, respectively.
As $x_1$ is fixed, the search space in $P'$
can be reduced to parameter fields which are constant in the $P'_{11}$ and $P'_{21}$ component,
say $P'_{11} = P'_{21} = 0,$
without increasing the functional value.
Let us denote such a reduced parameter field by
$\P'';$ i.e. $\P''(m) = (0, P'_{12}(m), P'_{13}(m); 0, P'_{22}(m), P'_{23}(m)).$ 
The remaining four  
entries of the reduced parameter field are estimated via
\begin{equation}\label{eq:horizontalProblemNewParameter}
\begin{split}
\hat \P'' = \argmin_{\P''} ~& \Bigg( \kappa \|\mbox{\boldmath{$\nabla$}}_{\d_1} \P'' \|_0 + \sum_{m=1}^n ( v'_1(m) - P'_{12}(m)m - P'_{13}(m) )^2
\\ & + ( v'_2(m) - P'_{22}(m)m - P'_{23}(m) )^2\Bigg).
\end{split}
\end{equation}
The crucial point 
is that 
the problem \eqref{eq:horizontalProblemNewParameter} 
can be solved exactly and efficiently 
and that
 $\hat \z(x_1, \cdot)$ in \eqref{eq:horizontalProblemNew} 
is recovered directly from the reduced parameter field $\hat\P''$ by
\begin{equation}
\begin{split}
    \hat z'_1(x_1, m) = \hat P'_{12}(m)m + \hat P'_{13}(m) \\
    \hat z'_2(x_1, m) = \hat P'_{22}(m)m + \hat P'_{23}(m) 
\end{split}
\end{equation}
without computing an optimal full parameter field $\hat \P'$
in \eqref{eq:horizontalProblemNew}.



We propose to solve problem \eqref{eq:horizontalProblemNewParameter} by dynamic programming. 
To that end, we cast it to a partitioning problem.
We denote by $\Ic$ a partition  of $\Nc= \{1, 2, \ldots, n\}$,
so that $\Ic$ consists of subsets of $\Nc$ such that $\cup_{I\in \Ic} = \Nc$ and $I \cap J = \emptyset$
whenever $I\neq J.$ Here,
we additionally require that each $I \in \Ic$ is a \enquote{discrete interval}; that is, $I$ is of the form $\{ l, l+1, \ldots, r\}.$
The minimum functional value in \eqref{eq:horizontalProblemNew} is equal to the minimum value of
the functional 
\begin{equation}\label{eq:1dPartitionProblem}
B(\Ic) =  \kappa  (|\Ic| - 1) +   \sum_{I \in \Ic} \sum_{t=1}^2 \min_{a,b \in \R} \sum_{p \in I}  (a p +b - v'_t(p))^2
\end{equation}
taken over all partitions $\Ic$ of $\Nc.$
(Note that the sum over $t=1, 2$ comes from expanding the Euclidean norm in $\R^2.$)
From an optimal partition $\hat \Ic$ which minimizes $B,$
the minimizer  $\hat \P''$ of \eqref{eq:horizontalProblemNewParameter} 
can be obtained by letting $\hat \P''$ on $I \in \hat\Ic$ 
 the (vectorial) affine linear parameters determined by 
\begin{equation}
(\hat P'_{t2}, \hat  P'_{t3}) = \argmin_{a,b} \sum_{p \in I}  (a p +b - v_t(x_1,p))^2,\quad\text{for }t=1, 2.
\end{equation}
It now remains to compute an optimal partition $\hat \Ic$ for  problem \eqref{eq:1dPartitionProblem}.
Our solver is based on the scheme presented in \cite{winkler2002smoothers,kleinberg2006algorithm,friedrich2008complexity} which we explain next.
We denote the optimal functional value 
for data given on the domain  $\{ 1, \ldots, r\}$ by
\begin{equation}
B_r^* = \min_{\Ic \text{ partition on }\{ 1, \ldots, r\}}  B(\Ic).
\end{equation}
It   satisfies the Bellman equation
\begin{equation}\label{eq:recurrencePenalized}
    B_r^* 
    = \min_{l= 1, \ldots, r} \left(B^*_{l-1} + \kappa + \sum_{t=1}^2\epsilon_{lrt}\right),
\end{equation}
where we let $B^*_0 = -\kappa$ and 
\begin{equation}\label{eq:epslrk}
    \epsilon_{lrt} = \min_{a,b\in \R} \sum_{p=l}^r (a p +b - v_t(x_1,p))^2.
\end{equation}
This reveals that $B_r^*$ can be computed from 
$B_{l-1}^*$ and $\epsilon_{lrt}$ for $l = 1, \ldots, r$ and for $t=1, 2.$
By the dynamic programming principle,
we successively compute  $B_1^*,$ $B_2^*,$ until we reach $B_n^*.$ 
As our primary interest is the partition $\Ic$ rather than the functional value,
we keep track of a corresponding optimal partition.
An economic way is to store at step $r$ the minimizing argument $l^*$ 
of \eqref{eq:recurrencePenalized}; 
see \cite{friedrich2008complexity} for a detailed description of that data structure.
We further note that the $\epsilon_{lrt}$
in \eqref{eq:epslrk} can be computed in $\Oc(1)$
by precomputation of the moments of the data in \eqref{eq:horizontalProblemNew};
see Appendix~\ref{app:approximationErrors} for a detailed description.
The worst case complexity of this algorithm is $\Oc(n^2)$,
 where $n$ is the number of elements in one line of the motion field.
Thus, we get the complexity $\Oc(N^{3/2})$, where $N$ denotes the number of pixels in the image.
 Since $n$ lines can be processed simultaneously, the complexity is $\Oc(N)$
  if $n$ processors are available.
 To further accelerate the computations, 
 we adopt the pruning strategy of~\cite{storath2014fast}.

So far, we have discussed the direction $\d_1.$
For the  directions $\d_2, \ldots, \d_K,$
we get intrinsically one-dimensional problems 
along the paths determined by the finite-difference vectors in a similar way.
More precisely, we solve the one-dimensional problems
of the form \eqref{eq:horizontalProblemNew} linewise along
vertical paths for $k =2$
and along diagonal and antidiagonal paths for $k=3,4,$ respectively.
The 1D subproblems in vertical direction have length $m$. Meanwhile, those in the vertical direction have varying lengths, because the number of pixels in a	
diagonal direction depends on its offset.

\section{Experimental results}
\label{sec:experiments}
\subsection{Large Displacements Model}
\label{sec:ldm}
Modern evaluation benchmarks often include large displacements. To cope with them, we extend the model described in Section \ref{sec:pwaff_of} by adopting the
approach described in \cite{Brox10,Weinzaepfel13,Revaud15}. It has become standard for variational motion estimation. This
amounts to adding a  term $\phi(\w,\m)$ to the model (\ref{eq:ms_para_field}), to promote similarity of the motion field $\w$ to the motion of a precomputed set of 
matched pixels $\m:\Lambda\subset\Omega\rightarrow\mathbb R^2$, defined on a sparse subset $\Lambda$ of the image grid.
This leads to
\begin{equation}
\label{eq:ldm}
\begin{split}
J=&\min_{\w,\P}~\rho(\w) + \gamma\phi(\w,\m) + \lambda \sum_{k=1}^K\alpha_k   
\| \mbox{\boldmath{$\nabla$}}_{\d_k} \P \|_0 \\
&\text{s.t. }\w(\x) = \P(\x)\bar{\x} 
\end{split}
\end{equation}
$\text{for all }\x \in \Omega$, where $\gamma>0$ is a balance parameter and $\phi(\w,\m)$ is defined by
\begin{equation}
\phi(\w,\m) = \sum_{\x\in\Omega} c(\x)\,\|\w(\x) - \m(\x)\|_1,
\end{equation}
where $c$ is the indicator function of $\Lambda$ defined by 
\begin{equation}
c(\x)=
\begin{cases}
1, &\x\in\Lambda\\
0, &\text{else.}
\end{cases}
\end{equation}
We compute the matches with the method described in \cite{Weinzaepfel13}, using the public code of the authors\footnote{\url{http://lear.inrialpes.fr/src/deepmatching/}}.
This new term has almost no impact on the computational cost in the optimization framework described in Section \ref{sec:pwaff_of}.
We introduce an additional splitting variable associated to $\phi$, which generates a new subproblem that is solved directly, like in Section \ref{sec:subprb_w}. We give
the detailed minimization steps in Appendix \ref{app:ldm}.

Finally, to overcome the restriction of small displacements of the linearized data term (\ref{eq:data_term}), the estimation is integrated in a standard
coarse-to-fine scheme \cite{Brox04}.


  \subsection{Implementation Details}

The parameters are optimized on a subset of 30\% of the {\it training} data set, both on  {\it MPI Sintel} and {\it Kitti}. The results of Table \ref{table:benchmarks} are obtained on the rest of the sequences.
To accelerate convergence, we increase the value of $\eta$ at each iteration in (\ref{eq:admm}). We start with the
initial value $\eta^{(0)}=0.01$ and we define the sequence $(\eta^{(i)})_{i\in\mathbb N}$ by a geometric evolution $\eta^{(i+1)}=\tau\, \eta^{(i)}$ with $\tau=1.1$.
As pre-processing, we apply Gaussian filtering with a variance of 0.9 to the input images to reduce the influence of noise.
We apply a weighted median filter as a post-processing at each scale of the coarse-to-fine scheme to remove outliers. 
The scale factor of the coarse-to-fine pyramid is set to 0.75.

The algorithm has been implemented in MATLAB, with a C++ implementation for the dynamic programming solver. The 1D piecewise-affine denoising subproblem (Section
\ref{sec:subpb_z}), which consumes most of the computational time, is naturally parallelizable. The reported runtime results have been obtained with a parallelization on
4 cores.

\subsection{Comparison Methods}

  \begin{table}[!t]
  \centering
 \caption{Comparison of AEP on benchmarks.}
 \small
 \begin{tabular}{lcc}
  \toprule
\multicolumn{3}{c}{ \bf\it MPI Sintel} \\
 \midrule
\multirow{2}{*}{ Method} & AEP   & AEP \\
 & Training set  & Test set \\
 \midrule
{\bf Ours}  & {\bf 2.27}  & {\bf 1.35} \\[2pt] 
Ours-TV  & 2.53 & - \\
 {\it DataFlow} \cite{Vogel13}  & 5.16   & - \\[2pt] 
{\it DeepFlow} \cite{Weinzaepfel13}  & 3.04  & 1.77 \\[2pt] 
{\it Classic+NL} \cite{Sun14}  & 5.22  & 3.77 \\[2pt] 
{\it PH-Flow} \cite{Yang15} & - & 1.71 \\[2pt] 
{\it FC-2Layers-FF}  \cite{Sun13}  & - & 3.05 \\
   \end{tabular}
~\\[2pt]
 \begin{tabular}{lcc}
  \toprule
\multicolumn{3}{c}{ \bf\it Kitti} \\
 \midrule
\multirow{2}{*}{ Method} & AEP   & AEP \\
 & Training set  & Test set \\
  \midrule
{\bf Ours}  & {\bf 1.29}  & 1.5 \\[2pt] 
Ours-TV  & 1.56 & - \\[2pt] 
 {\it DataFlow} \cite{Vogel13}  & 1.36   & 1.9 \\[2pt] 
{\it DeepFlow} \cite{Weinzaepfel13}  & 1.45  & 1.5 \\[2pt] 
{\it Classic+NL} \cite{Sun14} & 2.61  & 2.8 \\[2pt] 
{\it PH-Flow} \cite{Yang15}  & - & {\bf 1.3} \\[2pt] 
{\it NLTGV-Census}  \cite{Ranftl14}  & - & 1.6 \\ 
\bottomrule
 \end{tabular}
  \label{table:benchmarks}
  \end{table}

We want to focus on regularization while validating our piecewise-affine model. We
consider competing methods that are as close as possible to ours and compare our method with 1) the usual TV and TGV regularizations, and 2)  other
piecewise-parametric approaches. \\

\noindent{\bf TV-Based Methods}~
The method named {\it Classic++} is described in \cite{Sun14}. It uses the same data term as ours, with an anisotropic TV regularization but without the features
 described in Section \ref{sec:ldm}. 
 
 The $DeepFlow$ method \cite{Revaud15} differs from our formulation of Section \ref{sec:ldm} by an isotropic TV
 regularization instead of our piecewise-affine constraint and by a gradient conservation in addition to the intensity conservation (\ref{eq:data_term}).

Finally, we also consider the nonlocal extension of TV  described in \cite{Sun14} and named  {\it Classic+NL}.

To demonstrate the importance of the piecewise-affine model compared to TV
regularization, we create a method that we term Ours-TV by replacing the one-dimensional piecewise-affine constraint in (\ref{eq:ldm}) by a one-dimensional TV regularization. This leads to the optimization problem
\begin{equation}\label{eq:ours-TV}
J=\min_{\w}~ \left( \rho(\w) + \gamma\phi(\w,\m) + \lambda \sum_{k=1}^K\alpha_k \, TV_k(\w)\right), 
\end{equation}
where $TV_k$ applies TV regularization in the $k$th direction.
The minimization framework remains unchanged, except for the subproblems with respect to $\z_k$ in (\ref{eq:admm}). They
become TV-$\ell_2$ denoising problems, efficiently solvable with the taut-string algorithm \cite{condat2012direct}.  \\

\begin{figure*}[!t]
  \centering
   \begin{tabular}{m{2pt}m{175pt}@{\hspace{5pt}}m{50pt}@{\hspace{5pt}}m{175pt}@{\hspace{5pt}}m{50pt}}
& \includegraphics[height=75pt]{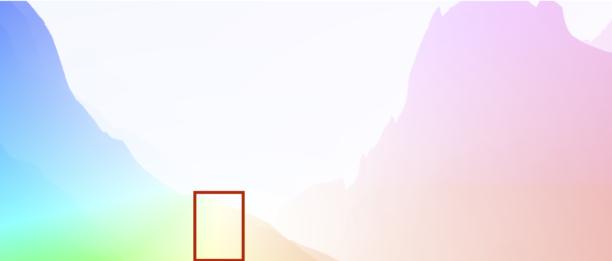} & \includegraphics[height=75pt]{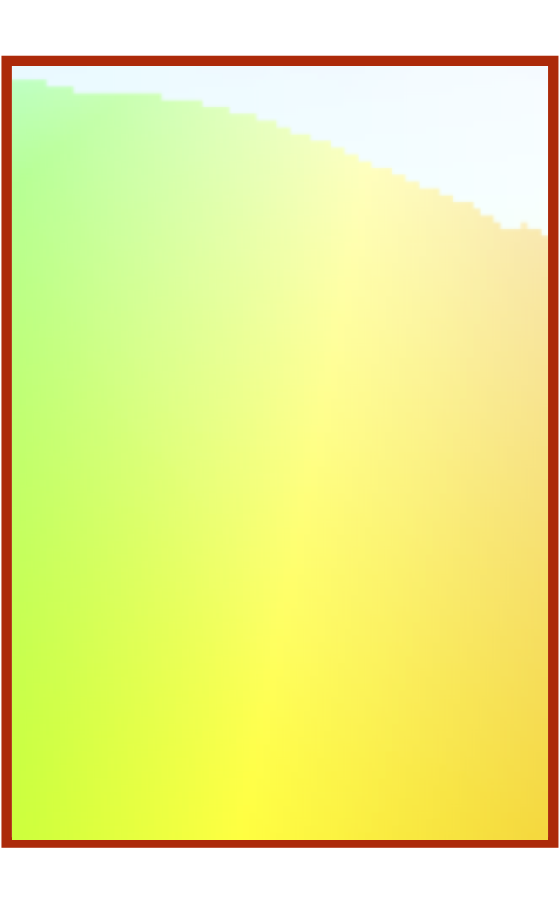} 
&\includegraphics[height=75pt]{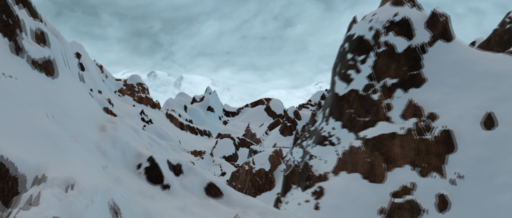} & \includegraphics[height=75pt]{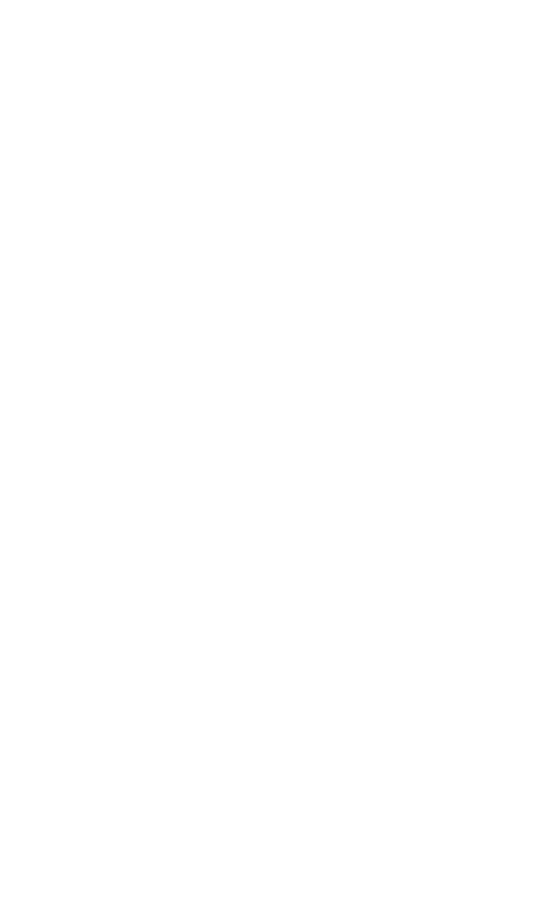}\\[-5pt] 
&\centering\mbox{Ground truth} &&   \centering\mbox{Overlay of the input images} & \\[15pt]	

\rotatebox{90}{Our method}&\includegraphics[height=75pt]{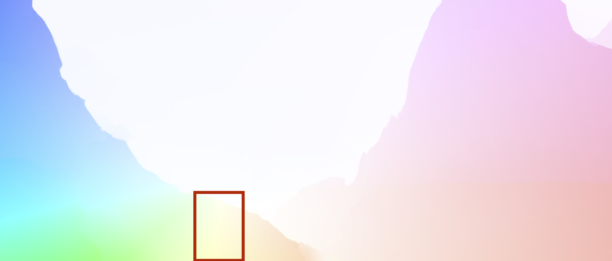} &
\includegraphics[height=75pt]{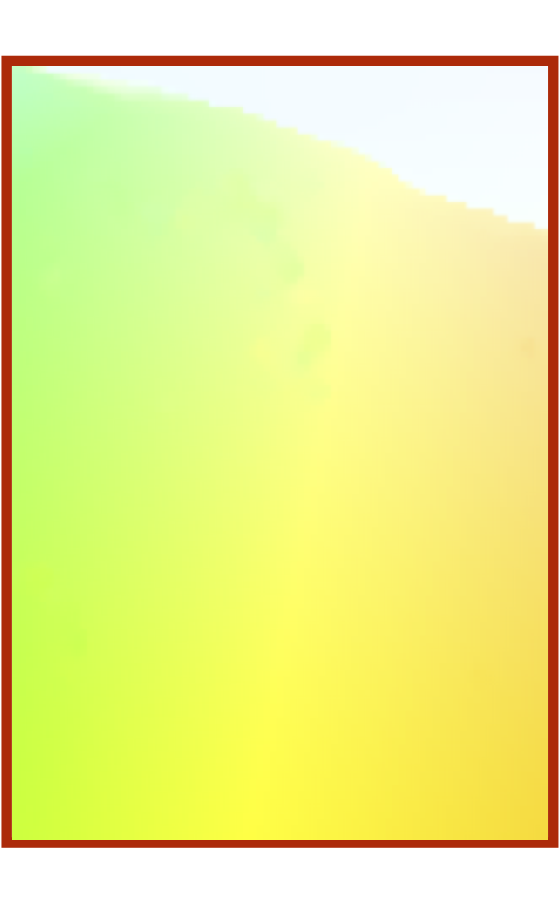}& \includegraphics[height=75pt]{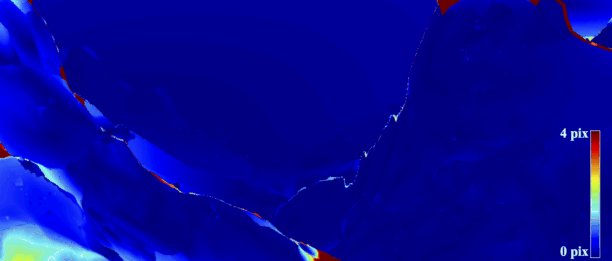} &
\includegraphics[height=75pt]{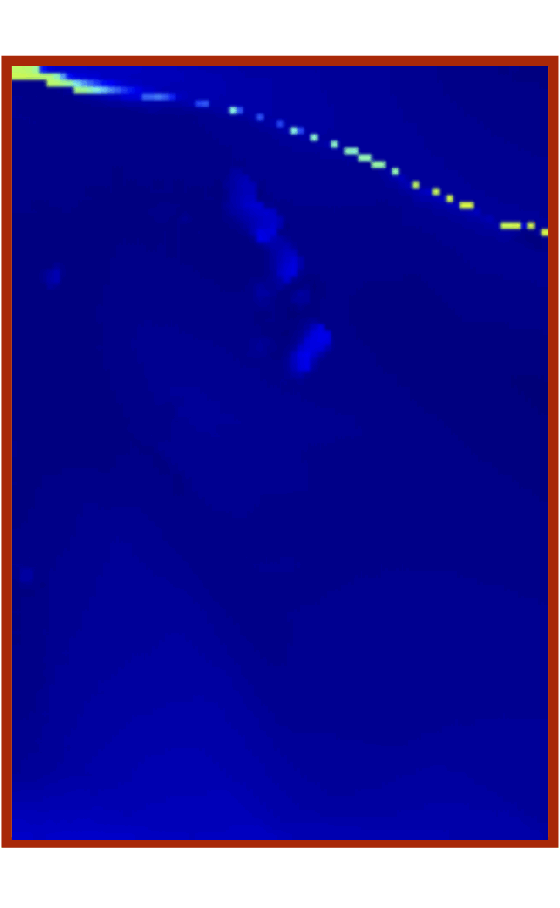}\\[-5pt] &
 && \centering\mbox{$\text{AEP}=0.15$} & \\[5pt]

 \rotatebox{90}{Ours-TV}& \includegraphics[height=75pt]{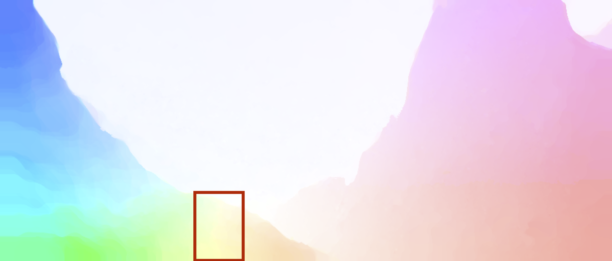}& \includegraphics[height=75pt]{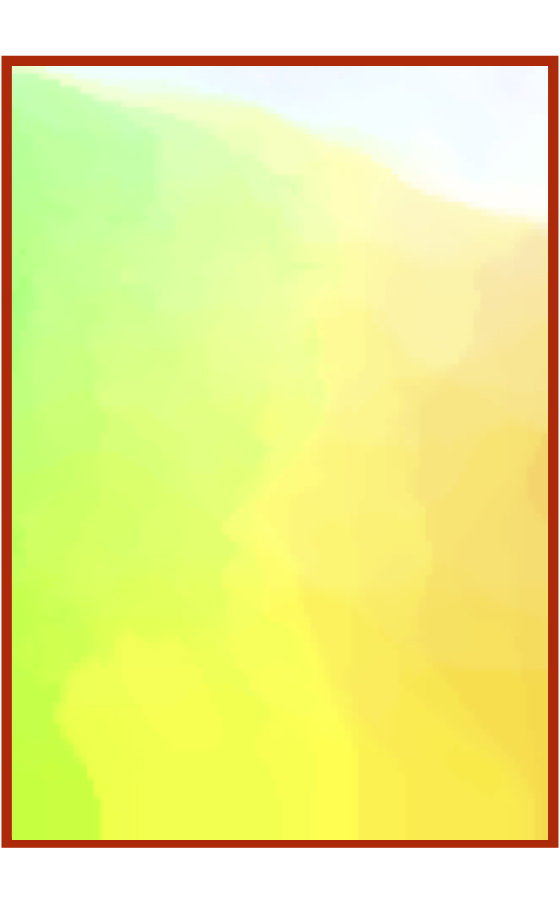} &
\includegraphics[height=75pt]{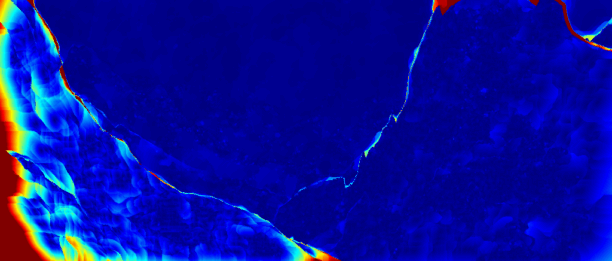} & \includegraphics[height=75pt]{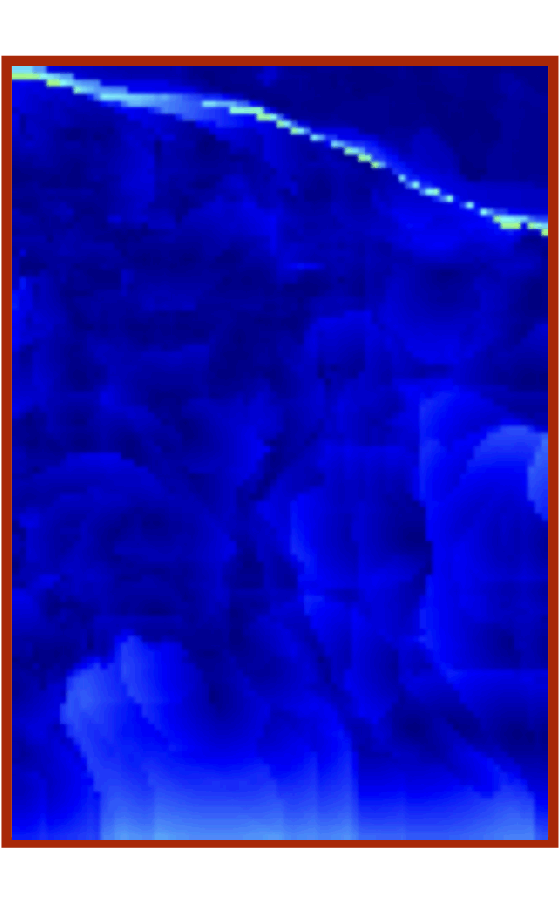}\\[-5pt]
& &&\centering \mbox{$\text{AEP}=0.26$} & \\[10pt]

\rotatebox{90}{\it Classic++}&\includegraphics[height=75pt]{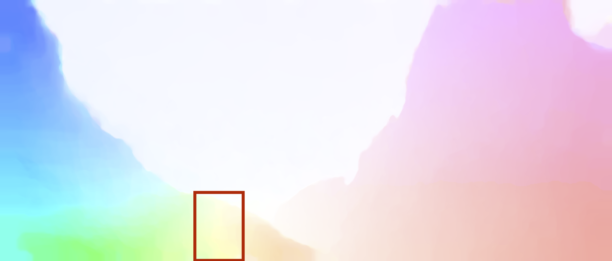} &
\includegraphics[height=75pt]{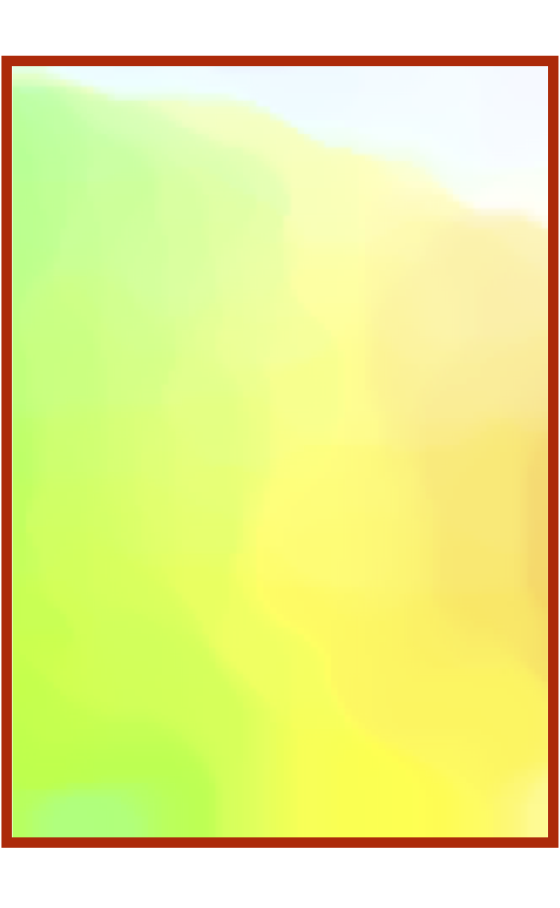}&
\includegraphics[height=75pt]{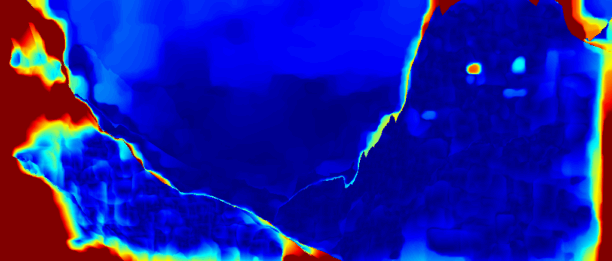} &
\includegraphics[height=75pt]{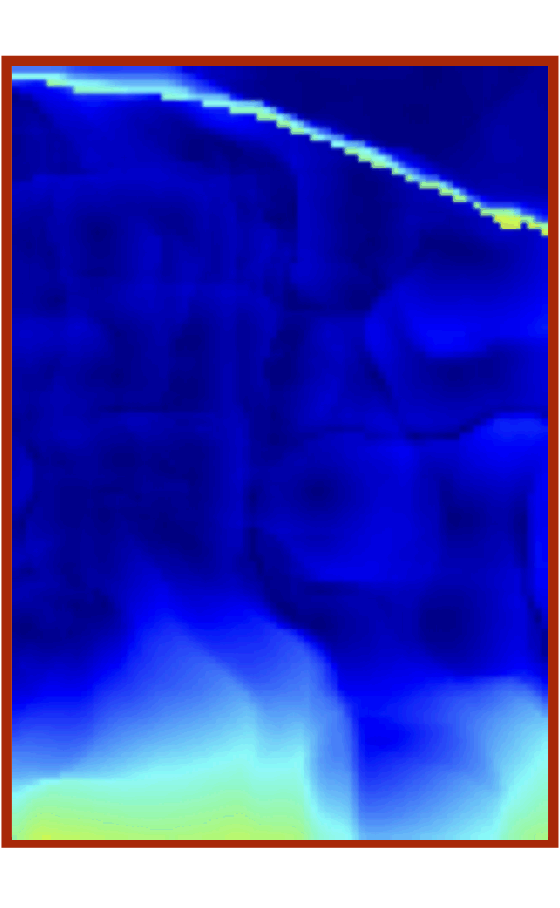}\\[-5pt]
 &  &&\centering \mbox{$\text{AEP}=0.99$} & \\[5pt]

\rotatebox{90}{\it Classic+NL}&\includegraphics[height=75pt]{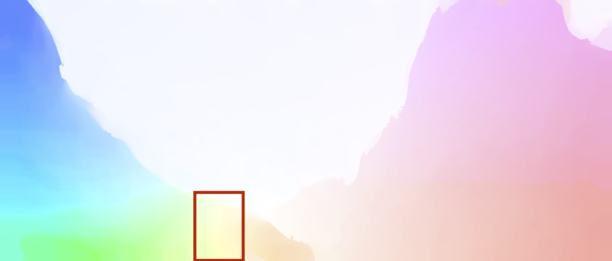} &
\includegraphics[height=75pt]{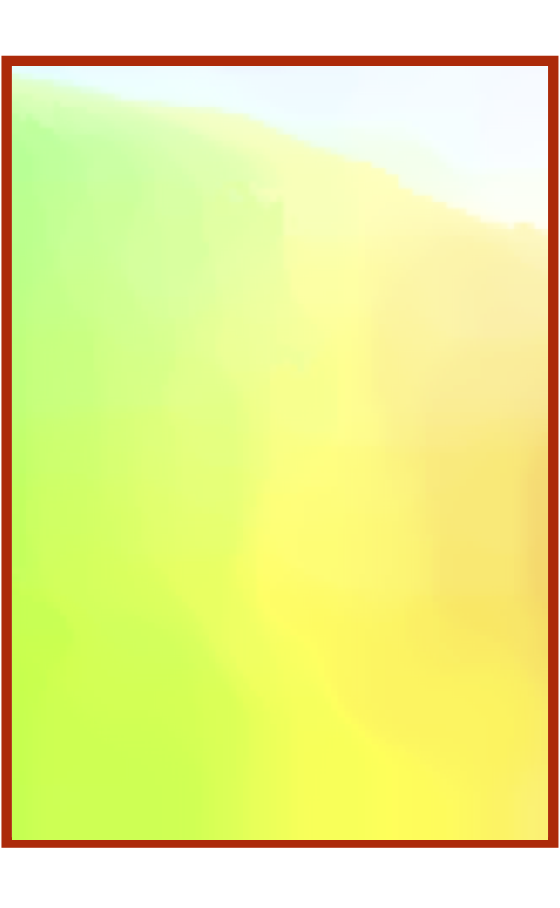}
& \includegraphics[height=75pt]{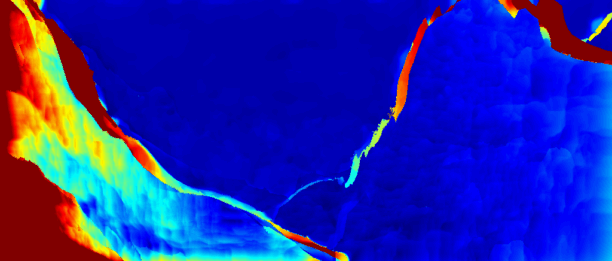} & \includegraphics[height=75pt]{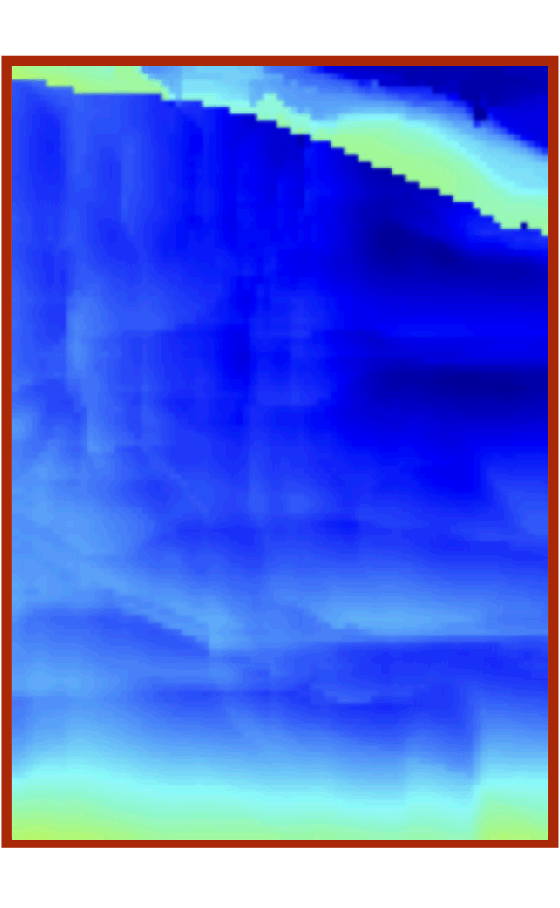}\\
&  && \centering \mbox{$\text{AEP}=0.96$} & \tabularnewline
   \end{tabular}
  \caption{Comparison of the TV motion fields (left column) and their associated error maps (right column). We provide a zoomed cutout of the motion
  field in each case.}
    \label{fig:TV}
  \end{figure*}

\noindent{\bf TGV-Based Methods}~
The second-order TGV regularization generalizes the TV approach and imposes a piecewise-affine form by introducing penalization of second derivatives with an $\ell_1$
norm \cite{Bredies10}. We consider the method $DataFlow$ described in \cite{Vogel13}. It uses TGV regularization with several advanced data terms. In our experiments, we
used the sum of absolute differences, which is a patch-based version of brightness constancy. We integrated $\phi$ from (\ref{eq:ldm}) in {\it DataFlow}, taking advantage
of the public code provided by the authors.

We also performed comparisons with the nonlocal version of TGV proposed in \cite{Ranftl14}, termed {\it NL-TGV}. However, the data term in {\it NL-TGV} is different from
ours.
It is based on the census transform, which provides  invariance to illumination changes.\\

\noindent{\bf Piecewise-Parametric Methods}~
The method termed {\it PH-Flow}   estimates
a piecewise homography model and is based on the formulation (\ref{eq:pwaff_mumsha}) with inter-piece regularization and graph cut optimization \cite{Yang15}.

We also perform comparisons with the method {\it FC-2Layers-FF} \cite{Sun13}, which is based on a layered representation and is not purely piecewise-parametric but allows
deviations from an affine model in each of the segmented region.
\\

We used the publicly available codes for {\it Classic++} and {\it Classic+NL}\footnote{\url{http://people.seas.harvard.edu/~dqsun/}},
$DeepFlow$\footnote{\url{http://lear.inrialpes.fr/src/deepflow/}}, and $DataFlow$\footnote{\url{http://github.com/vogechri/DataFlow/}}.

\begin{figure*}[!t]
  \centering
   \begin{tabular}{c@{\hspace{3pt}}c@{\hspace{3pt}}c}
\includegraphics[width=170pt]{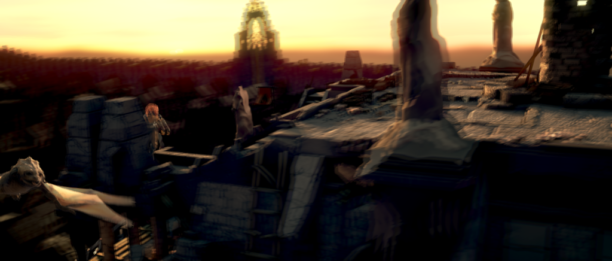}& \includegraphics[width=170pt]{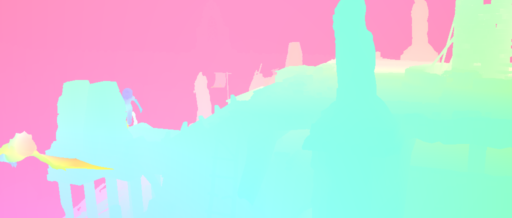} &
  \includegraphics[width=170pt]{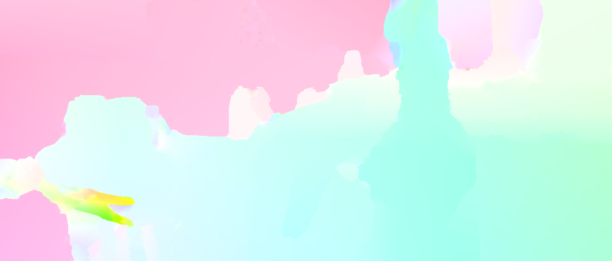}\\
  \mbox{Overlay of the input images} & \mbox{Ground truth} & \mbox{Our method} \\[3pt]
 \includegraphics[width=170pt]{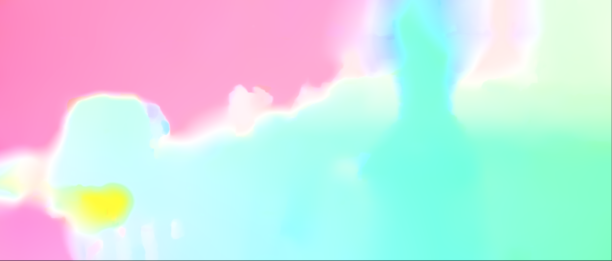}& \includegraphics[width=170pt]{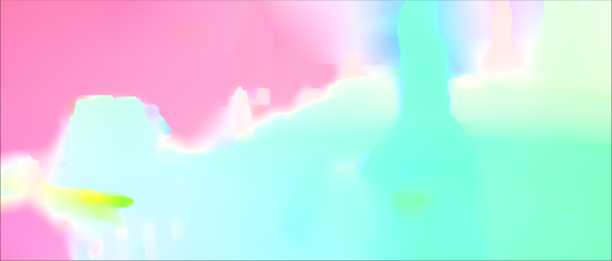} &
  \includegraphics[width=170pt]{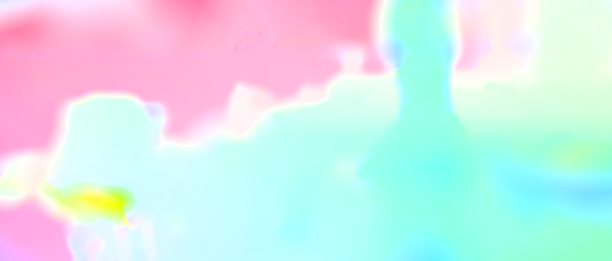}\\
  \mbox{\it TGV-Census} & \mbox{\it NLTGV-Census} & \mbox{\it DataFlow} \\
  {\small (extracted from \cite{Ranftl14})} & {\small (extracted from \cite{Ranftl14})} &  \\
   \end{tabular}
  \caption{Comparison of the TGV motion fields. The visualizations  of {\it TGV-Census} and
  {\it NLTGV-Census} are taken from \cite{Ranftl14}. The input images are from the $final$ version of the {\it MPI Sintel} dataset.}
    \label{fig:TGV}
  \end{figure*}

\subsection{Evaluation Datasets}
We validate our method on two reference benchmarks for motion estimation.

The {\it MPI Sintel} benchmark is composed of sequences extracted from a realistic animated movie. It contains $1\,064$ training sequences with available
ground truth and 564 test sequences used for blind evaluation \cite{Butler12}.
Each sequence has a {\it final} and a {\it clean} version. The {\it final} version introduces perturbations such as motion blur, defocus, or
atmospheric fog, which are not present in the {\it clean} version. These effects are handled by the data term or by specific estimation strategies.
To focus on the evaluation of the regularization, we used the {\it clean} dataset in our experiments.

The {\it Kitti} benchmark \cite{Geiger12} is composed of 193 training sequences and 193 test sequences, acquired in real outdoor conditions on a
platform installed on a moving car. A ground truth is provided only for half of the pixels. This benchmark is characterized by large illumination changes.

We compute the estimation accuracy with the endpoint error, defined at each pixel as the Euclidean distance between
the estimated motion vector and the ground truth. We report the averaged endpoint error (AEP) on the whole image. To isolate the impact of the regularization, all the
errors reported in this section have been computed in non-occluded regions.
Occlusion handling is a separate problem that requires dedicated techniques not discussed in this paper \cite{Fortun16,Ince08,Xiao06,Xu12}. 

The results on the two benchmarks are presented in Table \ref{table:benchmarks}, on the {\it training}
and {\it test} sets. We consider methods with public codes for the training set, and the ones with published results for the test set. Therefore, some methods are not
present in both categories. We did not report the result of {\it DataFlow} for the {\it test} set of {\it MPI Sintel} since the published results have been obtained
without the large displacement extension of Section \ref{sec:ldm}, which is decisive to obtain comparable results.

\subsection{TV and TGV regularization}
\label{sec:resTVTGV}

In Figure \ref{fig:TV}, we compare our result with  methods based on TV regularization, namely, Ours-TV, {\it Classic++}, and its nonlocal variant {\it Classic+NL}, in
the case of smooth variations of the motion field and small displacements.
We display the estimated motion field and the endpoint error maps. The TV regularization produces typical staircasing artifacts due to the piecewise constancy
of the solution. This effect is emphasized in the cutouts of Figure \ref{fig:TV}.
Our piecewise-affine approach does not produce staircasing and is much closer to the ground truth, both visually and in terms of AEP. We
also observe that the motion discontinuities are more accurately recovered with our approach.

In Figure \ref{fig:TGV}, we compare our method with the methods {\it TGV-Census}, {\it NLTGV-Census} and {\it DataFlow}, which are  based on TGV regularization. The
absence of staircasing of TGV comes at the price of some blurring artifacts in the result.
Even the nonlocal approach {\it NLTGV-Census}, which is specifically designed to reduce blurring, cannot solve completely the problem. In contrast, our method
combines a good restitution of affine displacements	 with a satisfactory recovery of sharp motion discontinuities.

\begin{figure*}[!t]
  \centering
   \begin{tabular}{m{2pt}m{125pt}@{\hspace{1pt}}m{125pt}@{\hspace{1pt}}|@{\hspace{1pt}}m{125pt}@{\hspace{1pt}}m{125pt}}
&\includegraphics[width=125pt]{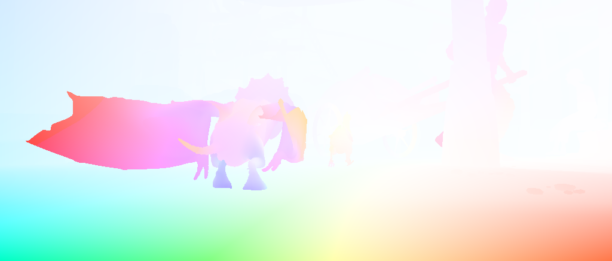}& \includegraphics[width=125pt]{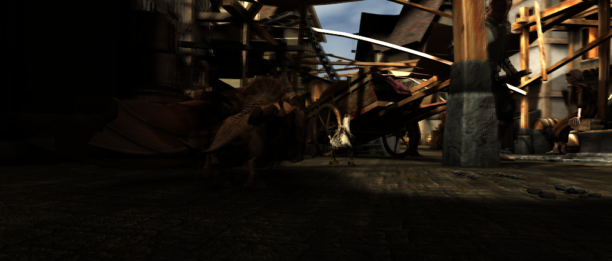}&  \includegraphics[width=125pt]{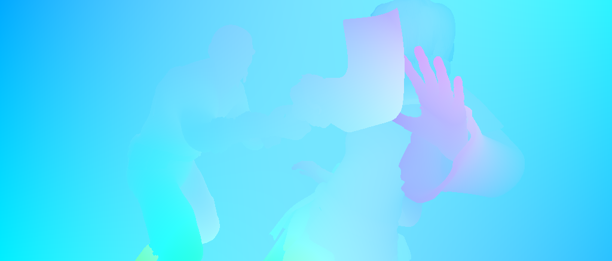}&
  \includegraphics[width=125pt]{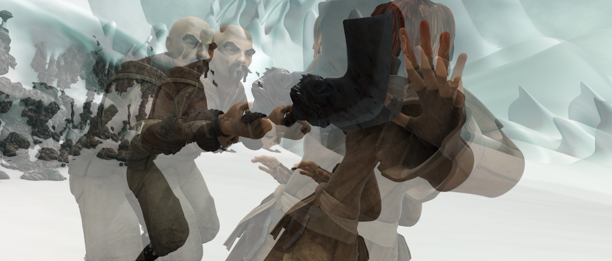}\tabularnewline[0pt]
&\centering \mbox{Ground truth} & \centering \mbox{Input images} & \centering \mbox{Ground truth} & \centering \mbox{Input Images}\tabularnewline[0pt]

 \rotatebox{90}{Our method}& \includegraphics[width=125pt]{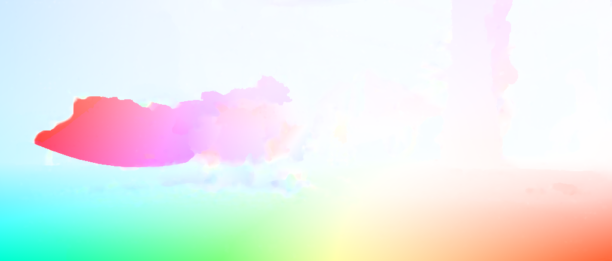}& \includegraphics[width=125pt]{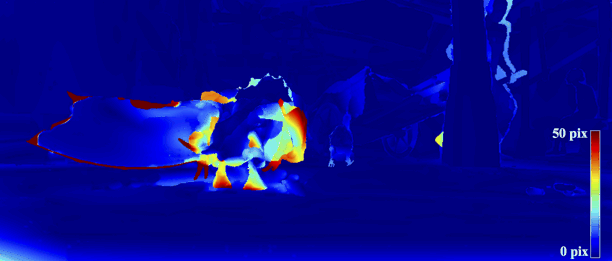} &
 \includegraphics[width=125pt]{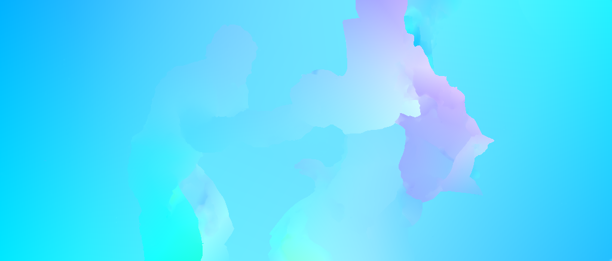} & \includegraphics[width=125pt]{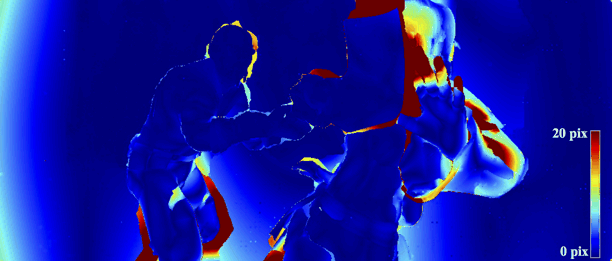} \tabularnewline
&&\centering\mbox{$\text{AEP}=1.55$}&  & \centering\mbox{$\text{AEP}=1.30$}\tabularnewline

\vspace{4pt}

\rotatebox{90}{Ours-TV}& \includegraphics[width=125pt]{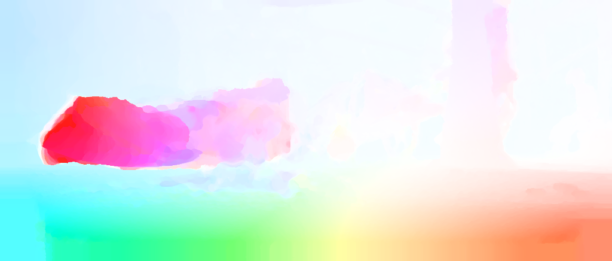}& \includegraphics[width=125pt]{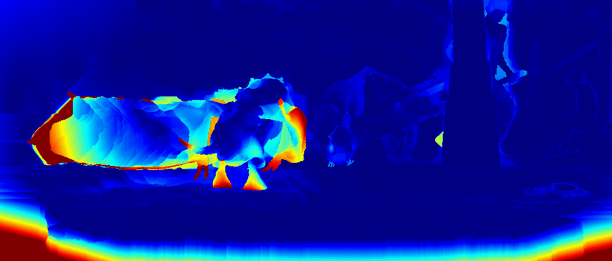} &
 \includegraphics[width=125pt]{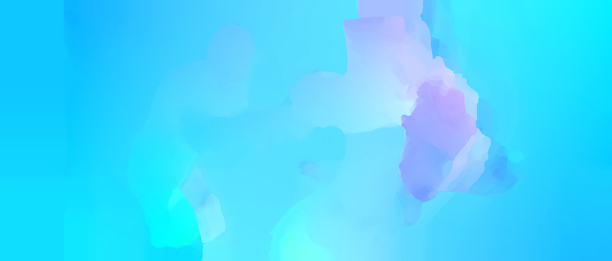} & \includegraphics[width=125pt]{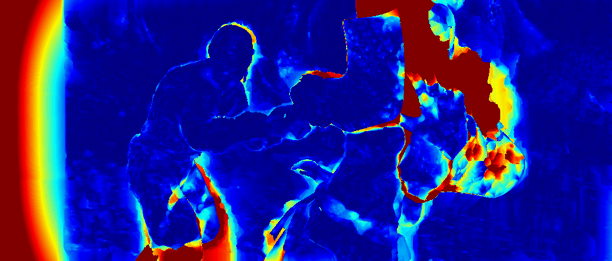} \tabularnewline
&&\centering\mbox{$\text{AEP}=2.02$}&  & \centering\mbox{$\text{AEP}=1.95$}\tabularnewline

\rotatebox{90}{{\it DeepFlow}}& \includegraphics[width=125pt]{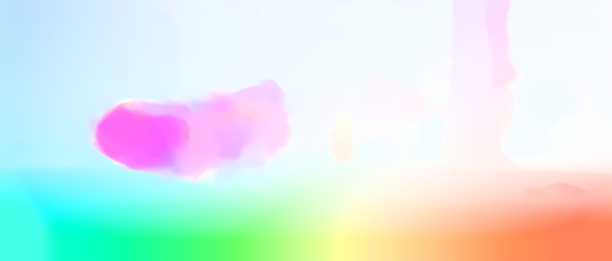}& \includegraphics[width=125pt]{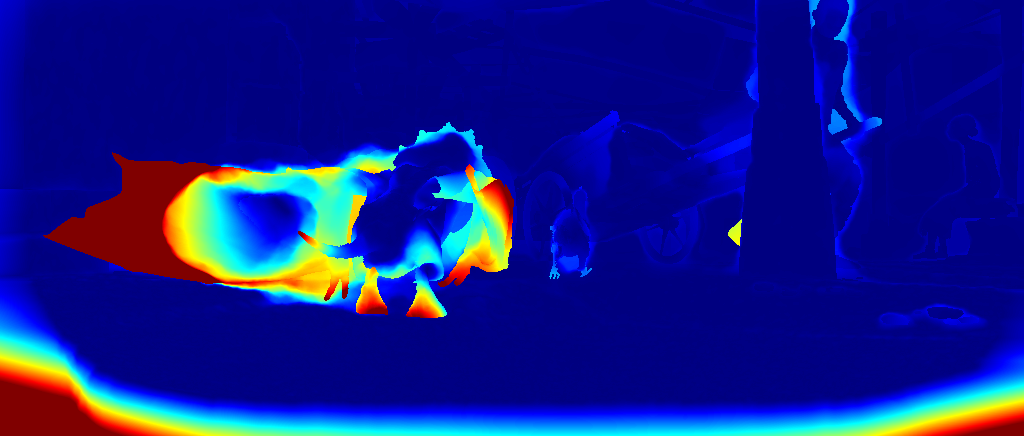} &
 \includegraphics[width=125pt]{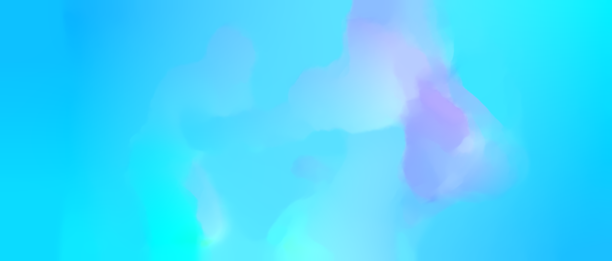} & \includegraphics[width=125pt]{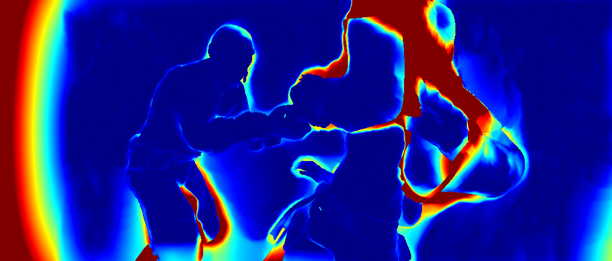} \tabularnewline
&&\centering\mbox{$\text{AEP}=3.98$}& & \centering\mbox{$\text{AEP}=2.01$}\tabularnewline

\rotatebox{90}{{\it DataFlow}}& \includegraphics[width=125pt]{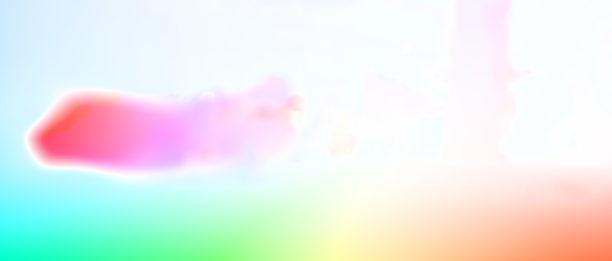}& \includegraphics[width=125pt]{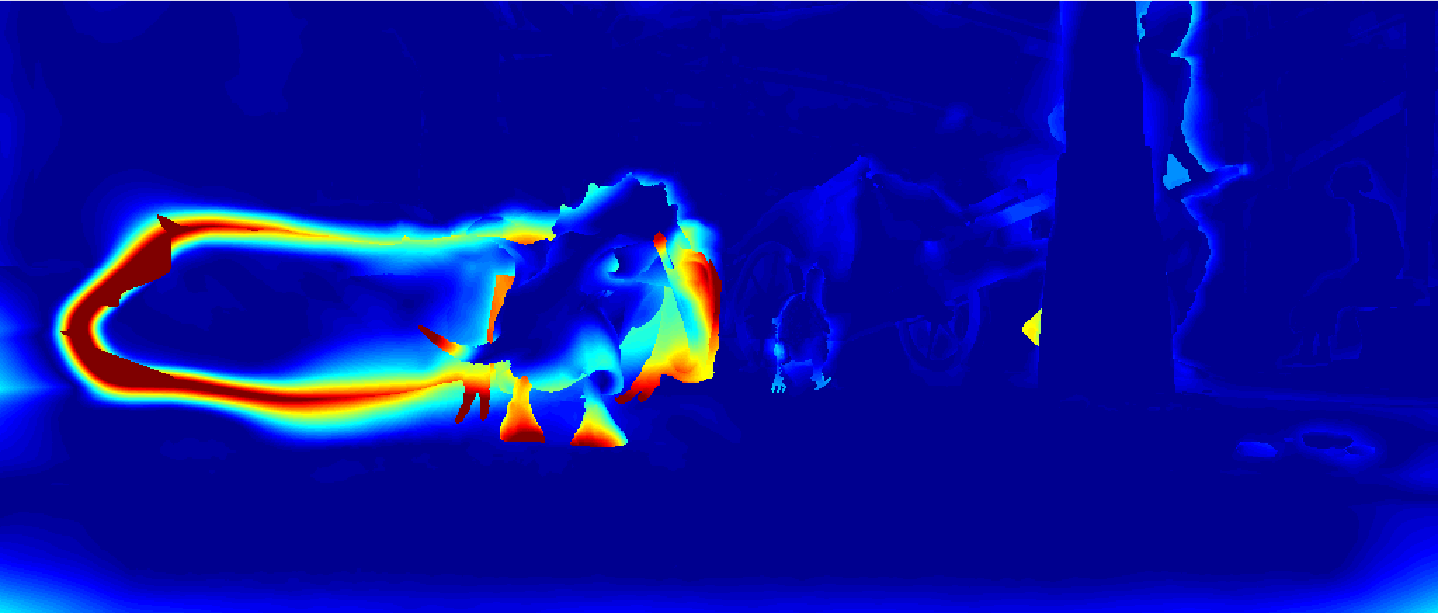} &
 \includegraphics[width=125pt]{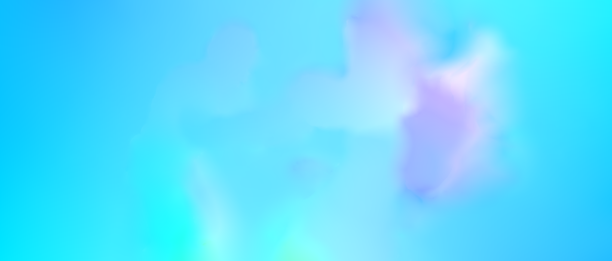} & \includegraphics[width=125pt]{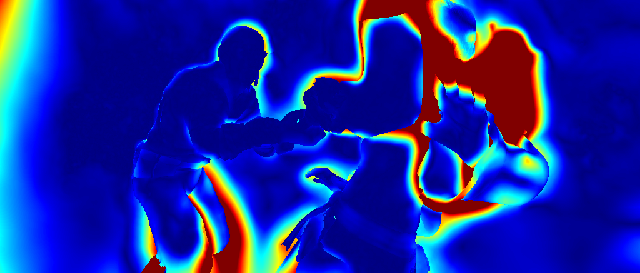} \tabularnewline
&&\centering\mbox{$\text{AEP}=2.34$}&  & \centering\mbox{$\text{AEP}=1.90$}\tabularnewline
   \end{tabular}
  \caption{Comparison of the motion fields (columns 1 and 3) and their associate error maps (columns 2 and 4) on the sequences {\it market\_4} (columns 1 and 2) and
  {\it ambush\_2} (columns 3 and 4) of the {\it MPI Sintel} benchmark.}
    \label{fig:visu}
  \end{figure*}

In Figure \ref{fig:visu}, we compare the results of our method, Ours-TV, $DeepFlow$ and $DataFlow$. We recall that the essential difference between these  methods is
only the regularization strategy. We observe that the sharpness of discontinuities is always better
preserved in our results compared to $DeepFlow$ and $DataFlow$. Generally, the global shapes of moving objects are more accurately delineated by our method. We also
observe staircasing artifacts in the results of Ours-TV. Altogether, the best AEP is achieved by our method. Note that large errors at image
borders are due to occlusions and are not taken into account in the computation of the AEP.

These qualitative observations are confirmed by the better results of our approach in Table \ref{table:benchmarks} compared to the methods with a similar framework but
different regularization: Ours-TV, {\it DeepFlow}, {\it DataFlow}, {\it Classic+NL}, and {\it NLTGV-Census}. 

\subsection{Piecewise-Parametric Methods}
\begin{figure}[!t]
  \centering
   \begin{tabular}{m{2pt}m{115pt}@{\hspace{1pt}}m{115pt}}
\rotatebox{90}{Input images} &\includegraphics[width=115pt]{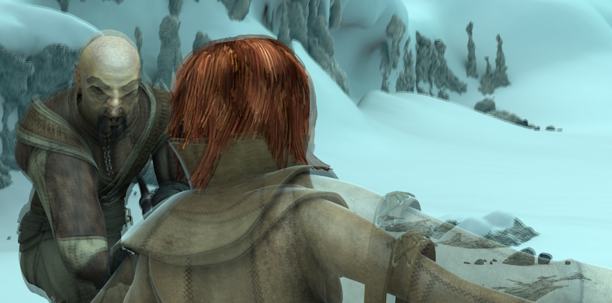}& \includegraphics[width=115pt]{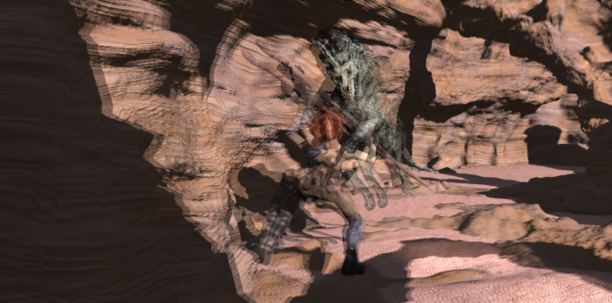}\tabularnewline

\rotatebox{90}{Ground truth	}&\includegraphics[width=115pt]{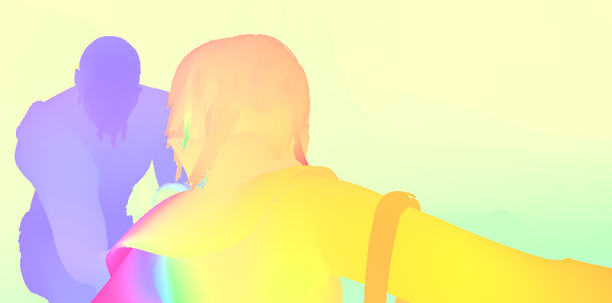}& \includegraphics[width=115pt]{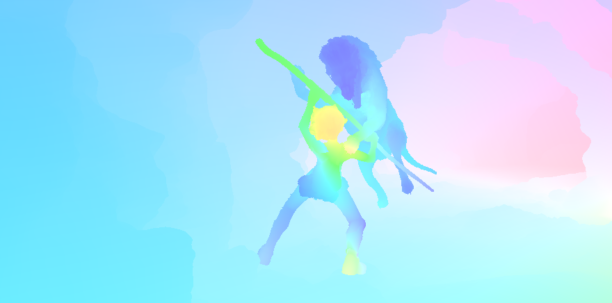} \tabularnewline

\rotatebox{90}{Our method}&\includegraphics[width=115pt]{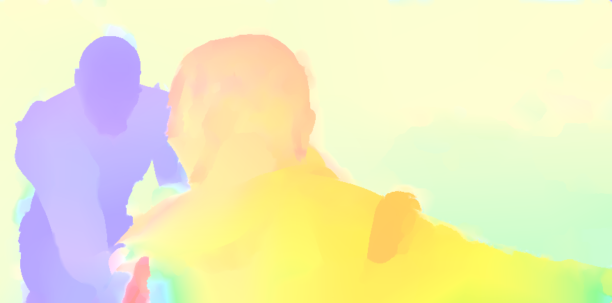}& \includegraphics[width=115pt]{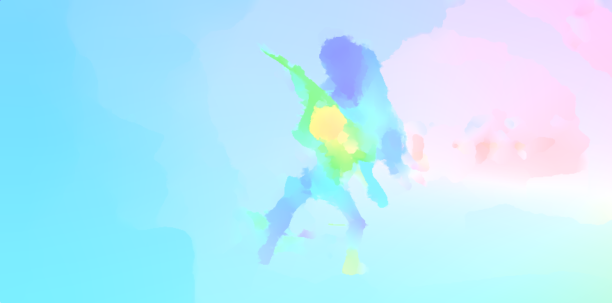}\tabularnewline
&\centering\mbox{$\text{AEP} = 3.60$} & \centering\mbox{$\text{AEP} = 0.34$}\tabularnewline

\rotatebox{90}{\it PH-Flow}&\includegraphics[width=115pt]{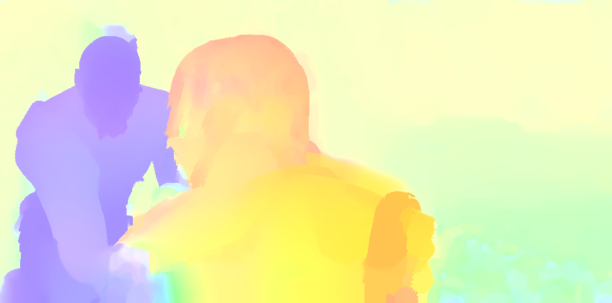}& \includegraphics[width=115pt]{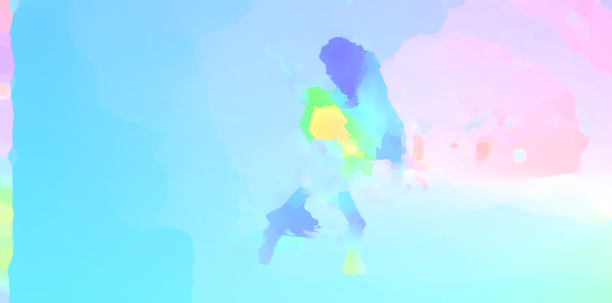}\tabularnewline
&\centering\mbox{$\text{AEP} = 4.11$} & \centering\mbox{$\text{AEP} = 0.47$}\tabularnewline

\rotatebox{90}{\it FC-2Layers-FF}&\includegraphics[width=115pt]{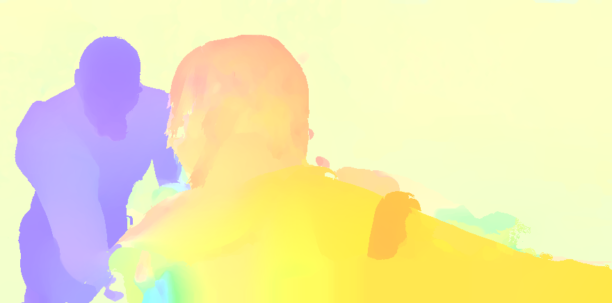}&
\includegraphics[width=115pt]{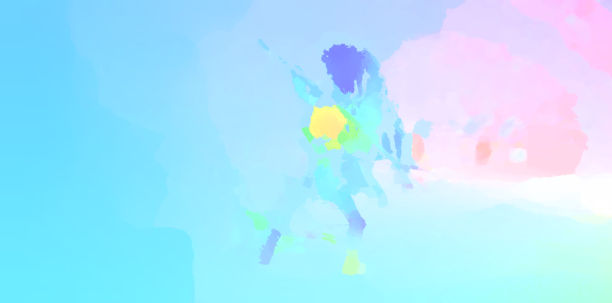}\tabularnewline
 &\centering\mbox{$\text{AEP} = 7.99$} & \centering\mbox{$\text{AEP} = 0.56$}\tabularnewline

   \end{tabular}
  \caption{Comparisons between the results of our method, {\it PH-Flow} \cite{Yang15}, and {\it FC-2Layers-FF} \cite{Sun13} on images from the {\it test} sequence  of the {\it MPI Sintel} benchmark.}
    \label{fig:piecewise_param}
  \end{figure}

In Figure \ref{fig:piecewise_param}, we show visual comparisons between our method, {\it PH-Flow}, and {\it FC-2Layers-FF}. On these examples, our method is able to
retrieve more details and to delineate motion discontinuities more accurately.
The numerical results of Table \ref{table:benchmarks} show a clear advantage of our method on {\it MPI Sintel}. On the {\it Kitti} benchmark, our method is close
to the best performing method {\it PH-Flow}.

Besides these numerical results, our method has four main practical advantages: (i) it is initialization-free, contrarily to  other piecewise-parametric motion
estimation methods; (ii) it is refinement-free, while the results of {\it PH-Flow} are obtained after refinement with the method {\it Classic+NL} dedicated to small
displacements; (iii) it is fast, taking  around 3 minutes on the non-downsampled
images of the {\it Kitti} benchmark, while  the reported computation time of {\it PH-FLow}  is 15 minutes on the same benchmark, despite a downsampling by a factor of 2
in \cite{Yang15}; (iv) its  running time is completely independent of the complexity of the flow field, while the running time of \cite{Yang15} is
depends highly on the complexity of the motion field due to its need for an explicit motion segmentation.

\footnotetext{We reproduce the images publicly available on the website of the benchmark. We also give the AEP associated to
  each motion field.}

\subsection{Reconstitution of Piecewise-Affine Edges}
\begin{figure*}[!t]
  \centering
   \begin{tabular}{m{3pt}m{160pt}@{\hspace{5pt}}m{160pt}@{\hspace{5pt}}m{160pt}}
 \rotatebox{90}{Input images} &\includegraphics[width=160pt]{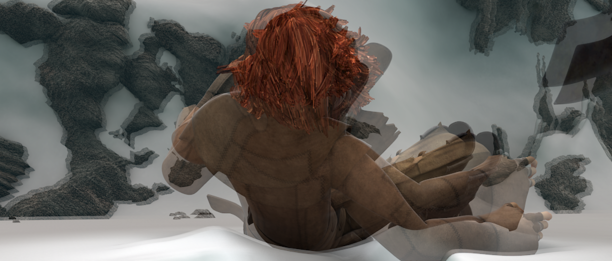}& \includegraphics[width=160pt]{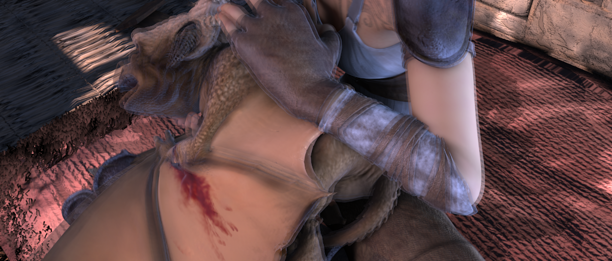} &
  \includegraphics[width=160pt]{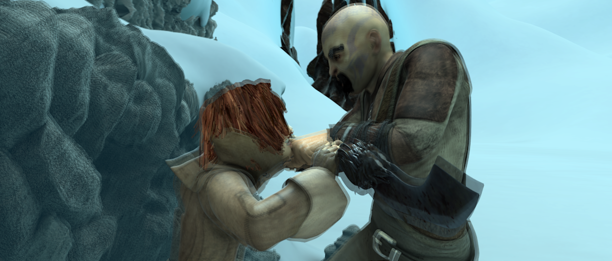}\tabularnewline
 \rotatebox{90}{Ground truth} &\includegraphics[width=160pt]{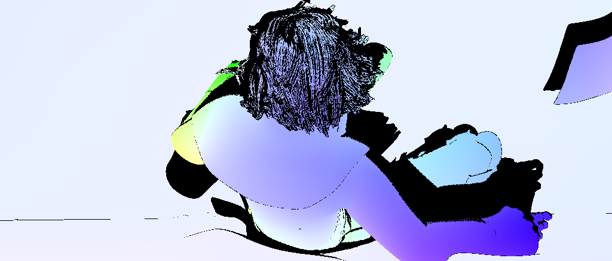}& \includegraphics[width=160pt]{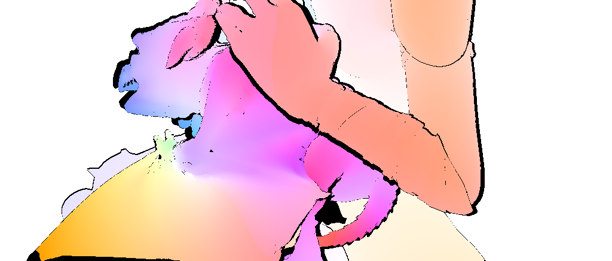} &
  \includegraphics[width=160pt]{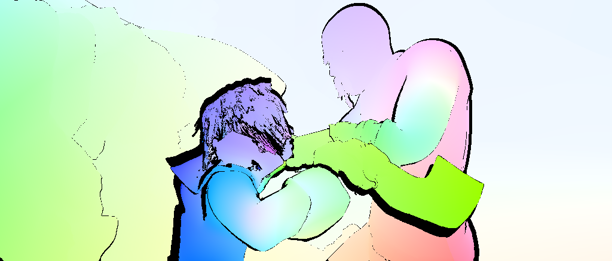}\tabularnewline
 \rotatebox{90}{Motion field}  & \includegraphics[width=160pt]{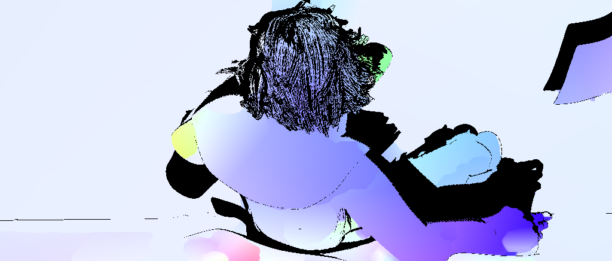}& \includegraphics[width=160pt]{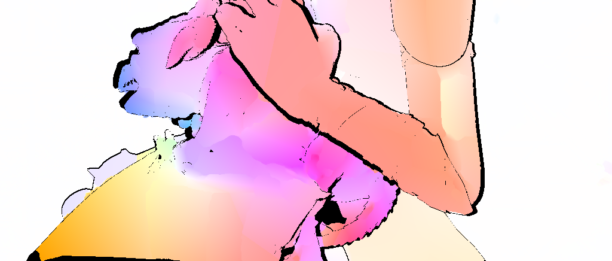} &
  \includegraphics[width=160pt]{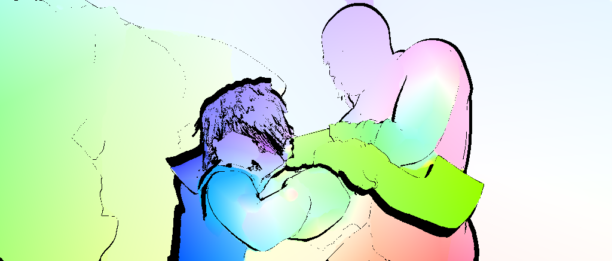}\\
 \rotatebox{90}{Motion edges} & \includegraphics[width=160pt]{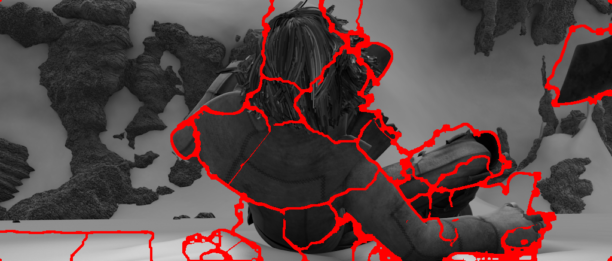}& \includegraphics[width=160pt]{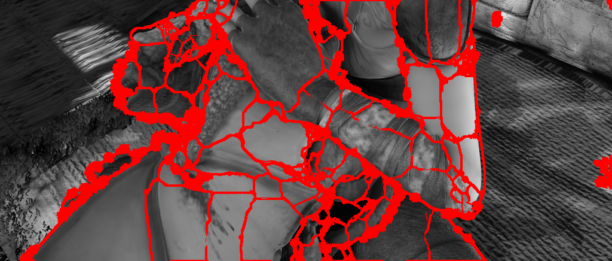} &
  \includegraphics[width=160pt]{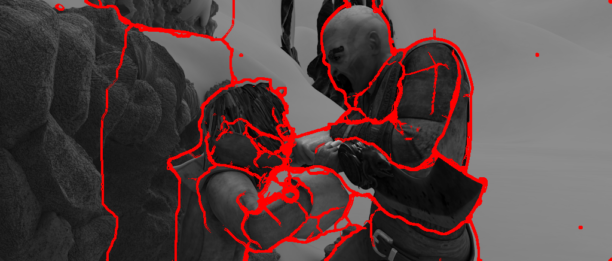}\\
   \end{tabular}
  \caption{Reconstruction of motion edges (last row) from the motion field estimated with our method (third row) on examples of the {\it MPI Sintel} dataset. Black pixels
  in the motion fields represent occluded region.}
    \label{fig:edges_sintel}
  \end{figure*}

To illustrate the piecewise-affine form of the estimated motion fields, we show in Figure \ref{fig:edges_sintel} a reconstruction of motion
edges obtained by thresholding the magnitude of spatial derivatives of the motion field.
When the scene is composed of a few moving parts 
undergoing simple deformations, the image domain is divided in a few meaningful regions. When the motion is more complex, our method  decomposes the motion field
in smaller pieces. A crucial aspect of our method is that this increasing complexity has no effect on the computational cost.

\section{Conclusion}  
We have proposed a new method to estimate piecewise-affine motion fields. 
In contrast to related methods, 
our approach does not rely on explicit segmentation
but  directly estimates a piecewise-affine motion field.
Key steps in the derivation 
are the specific formulation of the energy functional
as a constrained optimization problem
and the decomposition into tractable subproblems by an alternatig direction method of multipliers strategy.
Then, these subproblems are cast
to (non-convex) univariate piecewise-affine problems. A crucial ingredient of our method is that we are able to solve them exactly and efficiently. 
Our method overcomes the two main limitations of previous piecewise-parametric approaches, namely, sensitivity to initialization and computational cost.
Yet, our experiments show that it is competitive in terms of quality.
Further, they suggest that the piecewise affine model improves upon total variation and total generalized variation regularizations when using
similar data terms.
The versatility of our proximal splitting strategy lets extensions of the method to new data terms  be easily implemented.
Thus, the proposed approach can serve as a general regularization 
framework for motion estimation.

\section*{Acknowledgement}
This work was supported by the German Research Foundation DFG under Grant STO1126/2-1 and Grant WE5886/4-1, and by the European Research Council under Grant 692726 (H2020-ERC Project GlobalBioIm).

\appendices
\section{Calculation of the Approximation Errors}\label{app:approximationErrors}
We describe how to efficiently compute the  approximation errors $\epsilon_{lrt}$ required in \eqref{eq:recurrencePenalized}.
Let $g \in \R^{n\times 2}.$ (This corresponds to   $g_{pt} = v_t(x_1,p).$ for all $p, t$ in \eqref{eq:recurrencePenalized}.)
Taking the derivative of the right-hand side of (\ref{eq:epslrk}) with respect to $a,b$ yields
the optimality conditions
\begin{align}
	&\sum_{p=l}^r w_p (a_{lrt} p + b_{lrt} - g_{pt}) p = 0 \nonumber\\
	&\sum_{p=l}^r w_p (a_{lrt} p + b_{lrt} - g_{pt})  = 0.
\end{align}
This linear system can be rewritten as 
\begin{align}
	& a_k E_{lr} + b_k G_{lr}  =  I_{lrt}, \nonumber\\
	& a_k G_{lr} + b_k H_{lr}    =  J_{lrt},
\end{align}
with the auxiliary quantities
\begin{align}
    &E_{lr} = \sum_{p=l}^r w_p  p^2, \quad
	 G_{lr} = \sum_{p=l}^r w_p  p,  \quad
	 H_{lr} = \sum_{p=l}^r w_p,  \\
	&I_{lrk} = \sum_{p=l}^r w_p g_{pt} p, \quad
	J_{lrk} = \sum_{p=l}^r w_p g_{pt}.
\end{align}
The solutions $a^*_{lrk}$ and $b^*_{lrk}$ are given by
\begin{equation}
	a^*_{lrt} = \frac{I_{lr} H_{lr} - G_{lr}J_{lrk}}{E_{lr}H_{lr} - G_{lr}^2}
\end{equation}
and
\begin{equation}
	b^*_{lrt} = \frac{E_{lr}J_{lr} - I_{lrk}G_{lr}}{E_{lr}H_{lr} - G_{lr}^2}.
\end{equation}
Plugging this into \eqref{eq:epslrk} gives us
\begin{equation}
	\epsilon_{lrk} = 
	\frac{J_{lr}^2 E_{lr} - 2 G_{lr} J_{lrt} I_{lrt} + H_{lr} I_{lrt}^2 + G_{lr}^2 K_{lrt} - H_{lr} E_{lr} K_{lrt} }{G_{lr}^2 - H_{lr}E_{lr}}
\end{equation}
where $K_{lrt} = \sum_{p=l}^r w_p g_{pt}^2.$
Note that the involved sums can be computed efficiently 
by utilizing precomputations of moments.
For example, $E_{lr}$ can be computed via
$E_{lr} = E'_{r} - E'_{l-1}$
where $E'_{t} = \sum_{p=1}^t w_p p^2.$
So $E_{lr}$ can be computed in $O(1)$
if the vector $E'$ is precomputed.
$E'$ in turn can be computed in $O(n).$ 
For the other summations, analogous schemes are applied.

\section{Optimization for the Large Displacement Model}
\label{app:ldm}

To solve the minimization problem (\ref{eq:ldm}) for the extended model, we follow the splitting scheme used for (\ref{eq:constrained_min}). We introduce an
additional splitting variable associated to the term $\phi$. Accordingly,  (\ref{eq:ldm}) rewrites 
\begin{eqnarray}
 \label{eq:constrained_min_ldm}
&\,&J=\min_{\w,\u, \P_1, \ldots, \P_K}\left( \rho(\w) + \gamma\, \phi(\u,\m) + \lambda\sum_{k=1}^K \alpha_k \| \mbox{\boldmath{$\nabla$}}_{d_k} \P_k \|_0
\right)\nonumber\\
&\,& \text{ s.t. }
 \w(\x)=\z_k(\x),\nonumber \\
 &\,&  \phantom{\text{ s.t. }}\u(\x)=\w(\x),\label{eq:constrained_form_2} \\
 &\,& \phantom{\text{ s.t. }}
 \z_k(\x)=\P_k(\x) \bar\x ,~ \forall \x\in\Omega,~ \forall  k = 1,\ldots,K\nonumber.
\end{eqnarray}

The augmented Lagrangian associated to (\ref{eq:constrained_min_ldm}) is then
\begin{equation}
    \begin{split}
\Ac_{\eta_1,\eta_2}&(\w,\u,\{\P_k\}_k,\{\z_k\}_k,\{\mbox{\boldmath{$\upmu$}}_k\}_k,\mbox{\boldmath{$\upxi$}}) =  \\
&\rho(\w) + \gamma\,\phi(\u,\m) + \lambda\sum_{k=1}^K \alpha_k \| \mbox{\boldmath{$\nabla$}}_{\d_k} \P_k \|_0 \\
&+ \frac{\eta_1}{2}\sum_{k=1}^K\sum_{\x\in\Omega}\left\|\w(\x)-\z_k(\x)+\frac{\mbox{\boldmath{$\upmu$}}_k(\x)}{\eta_1}\right\|_2^2- \frac{1}{2\eta_1}\left\|\mbox{\boldmath{$\upmu$}}_k(\x)\right\|_2^2\\
&+ \frac{\eta_2}{2}\sum_{\x\in\Omega}\left\|\u(\x)-\w(\x)+\frac{\mbox{\boldmath{$\upxi$}}(\x)}{\eta_2}\right\|_2^2 - \frac{1}{2\eta_2}\left\|\mbox{\boldmath{$\upxi$}}(\x)\right\|_2^2\\[1ex] 
\text{ s.t. }& \z_k(\x)=\P_k(\x) \bar\x ,~ \forall \x\in\Omega,~ \forall  k\in\{1,\ldots,K\}.
 \end{split}
\end{equation}

Similarly to the update scheme (\ref{eq:admm}), the ADMM steps involve minimizing
$\Ac_{\eta_1,\eta_2}(\w,\u,\{\P_k\}_k,\{\z_k\}_k,\{\mbox{\boldmath{$\upmu$}}_k\}_k,\mbox{\boldmath{$\upxi$}})$ with respect to $\w$, $\z_k$, and $\u$. The minimization problem with respect to $\z_k$ is the same as in Section \ref{sec:subpb_z}. We detail now	 the updates of $\w$ and $\u$, which are very similar
to the description of Section \ref{sec:subprb_w}.

\paragraph{Update of $\w$} ~\\
The minimization w.r.t $u$ in can be rewritten
\begin{eqnarray}
\min_\w \rho(\w) ~+~ \frac{\eta_1 K + \eta_2}{2}\sum_{\x\in\Omega}\left(\w(\x)-\t(\x)\right)^2,
\end{eqnarray}
where
\begin{eqnarray}
\t(\x) = \frac{1}{\eta_1 K+\eta_2}\left(\eta_1 K \	\r(\x) + \eta_2 \left(\u(\x)+\frac{\mbox{\boldmath{$\upxi$}}(\x)}{\eta_2}\right) \right). 
\end{eqnarray}
The problem is pointwise and admits a closed-form solution with the thresholding scheme
  \begin{equation}
  \w(\x) = \t(\x) + 
  \begin{cases}
  \frac{\mbox{\boldmath{$\nabla$}} I}{\eta_1 K}, & \rho_0(\t(\x))<-\frac{\|\mbox{\boldmath{$\nabla$}} I\|_2^2}{\eta_1 K}\\
  -\frac{\mbox{\boldmath{$\nabla$}} I}{\eta_1 K}, & \rho_0(\t(\x))>\frac{\|\mbox{\boldmath{$\nabla$}} I\|_2^2}{\eta_1 K}\\
  -\rho_0(\t(\x))\frac{\mbox{\boldmath{$\nabla$}} I}{\|\mbox{\boldmath{$\nabla$}} I\|_2^2}, & |\rho_0(\t(\x))|\leq\frac{\|\mbox{\boldmath{$\nabla$}} I\|_2^2}{\eta_1 K}.
  \end{cases}
  \end{equation}

\paragraph{Update of $\u$}~\\
The minimization w.r.t $u$ in writes
\begin{eqnarray}
J=\min_\u \left(\gamma\phi(\u,\m) + \frac{\eta_2}{2}\sum_{\x\in\Omega}\left\|\u(\x)-\v(\x)\right\|^2\right).
\end{eqnarray}
where $\v(\x)=\w(\x)-\frac{\mbox{\boldmath{$\upxi$}}(\x)}{\eta_2}$.

The problem is pointwise and admits a closed-form solution with the thresholding scheme
  \begin{equation}
  u_k(\x) = 
    \begin{cases}
    v_k(\x),&c(\x)=0,\\
  v_k(\x) + \frac{\gamma}{\eta_2}, &v_k(\x)-m_k(\x)<-\frac{\gamma}{\eta_2} \text{ and } c(\x)\neq 0\\
  v_k(\x) -\frac{\gamma}{\eta_2}, & v_k(\x)-m_k(\x)>\frac{\gamma}{\eta_2} \text{ and } c(\x)\neq 0\\
  m_k(\x), &|v_k(\x)-m_k(\x)|\leq\frac{\gamma}{\eta_2} \text{ and } c(\x)\neq 0.
  \end{cases}
  \end{equation}
where $k=\{1,2\}$ and we use the notations $\u(\x)=(u_1(\x),u_2(\x))$, $\v(\x)=(v_1(\x),v_2(\x))$, and $\m(\x)=(m_1(\x),m_2(\x))$.

{\footnotesize
\bibliographystyle{plain}
\bibliography{refs}
}

\end{document}